\documentclass[acmsmall,screen]{acmart}

\usepackage{amsmath, bm}
\usepackage{algorithm}
\usepackage{booktabs}
\usepackage{algorithmic}
\usepackage{subfigure}
\usepackage{multirow}
\usepackage[switch]{lineno}

\AtBeginDocument{%
  \providecommand\BibTeX{{%
    \normalfont B\kern-0.5em{\scshape i\kern-0.25em b}\kern-0.8em\TeX}}}

\setcopyright{acmcopyright}
\copyrightyear{2022}
\acmYear{2022}
\acmDOI{10.1145/3532624}

\acmJournal{TKDD}
\acmVolume{1}
\acmNumber{1}
\acmArticle{1}
\acmMonth{1}

\begin{document}

\title{A Survey on Deep Hashing Methods}

\author{Xiao Luo}
\email{xiaoluo@pku.edu.cn}
\author{Haixin Wang}
\email{wang.hx@stu.pku.edu.cn}
\author{Daqing Wu}
\email{wudq@pku.edu.cn}
\affiliation{%
  \institution{Peking University}
  \city{Beijing}
  \country{China}
  \postcode{100871}
}


\author{Chong Chen}
\authornote{Corresponding authors: Chong Chen and Minghua Deng. The work was done when Xiao Luo and Daqing Wu interned in Damo Academy, Alibaba Group.}
\email{cheung.cc@alibaba-inc.com}
\affiliation{%
  \institution{Alibaba Group}
  \city{Hangzhou}
  \country{China}
  \postcode{310000}
}

\author{Minghua Deng}
\authornotemark[1]
\email{dengmh@math.pku.edu.cn}
\affiliation{%
  \institution{Peking University}
  \city{Beijing}
  \country{China}
  \postcode{100871}
}

\author{Jianqiang Huang}
\email{jianqiang.jqh@gmail.com}
\author{Xian-Sheng Hua}
\email{huaxiansheng@gmail.com}
\affiliation{%
  \institution{Alibaba Group}
  \city{Hangzhou}
  \country{China}
  \postcode{310000}
}

\renewcommand{\shortauthors}{Luo et al.}


\begin{abstract}
Nearest neighbor search aims to obtain the samples in the database with the smallest distances from them to the queries, which is a basic task in a range of fields, including computer vision and data mining. Hashing is one of the most widely used methods for its computational and storage efficiency. With the development of deep learning, deep hashing methods show more advantages than traditional methods. In this survey, we detailedly investigate current deep hashing algorithms including deep supervised hashing and deep unsupervised hashing. Specifically, we categorize deep supervised hashing methods into pairwise methods, ranking-based methods, pointwise methods as well as quantization according to how measuring the similarities of the learned hash codes. Moreover, deep unsupervised hashing is categorized into similarity reconstruction-based methods, pseudo-label-based methods and prediction-free self-supervised learning-based methods based on their semantic learning manners. We also introduce three related important topics including semi-supervised deep hashing, domain adaption deep hashing and multi-modal deep hashing. Meanwhile, we present some commonly used public datasets and the scheme to measure the performance of deep hashing algorithms. Finally, we discuss some potential research directions in conclusion.
\end{abstract}



\maketitle

\section{Introduction}
Nearest neighbor search is one of the most fundamental problems in various fields including computer vision and machine learning. Its purpose is to find the closest point from the dataset to the query based on a certain distance. 
Nevertheless, when it comes to high-dimensional and large-scale data, the time cost of accurately finding the sample closest to the query is substantial. To tackle the challenge, recent researchers have paid more attention to approximate nearest neighbor search because in most cases it can meet the search needs and significantly reduce the search complexity.

Hashing is one of the most widely used methods because it is very efficient in terms of computation and storage~\cite{cantini2021learning}. Its purpose is to convert the {high-dimensional features} vectors into low-dimensional hash codes, so that the hash codes of the similar objects are as close as possible, and the hash codes of dissimilar objects are as different as possible. The existing hashing methods consist of local sensitive hashing \cite{charikar2002similarity,indyk1998approximate} and learning to hash. 
The purpose of local sensitive hashing is to map the original data into several hash buckets. The closer the original distance between objects is, the greater the probability of falling in the same hash bucket. Through this mechanism, many algorithms based on locally sensitive hashing have been proposed \cite{broder1997resemblance,broder1997syntactic,dasgupta2011fast,datar2004locality,motwani2006lower,o2014optimal}, which show high superiority in both calculation and storage. However, in order to improve the recall rate of search, these methods usually need to build many different hash tables, so their applications on particularly large data sets are still limited.

Since local sensitive hashing is data-independent, researchers try to get high-quality hashing codes by learning good hash functions. As two pioneering methods, i.e., spectral hashing and semantic hashing, have been proposed~\cite{weiss2009spectral,salakhutdinov2009semantic}, learning to hash has sparked considerable academic interest in both machine learning and data mining.
With the development of deep learning \cite{lecun2015deep}, getting hash codes through deep learning gets more and more attention for two reasons. The first reason is that the powerful representation capabilities of deep learning can learn very complex hash functions. The second reason is that deep learning can achieve end-to-end hashing codes, which is very useful in many applications. {In this survey, we mainly focus on deep supervised hashing methods and deep unsupervised hashing methods, which are two mainstreams in hashing research. 
Moreover, three related important topics including semi-supervised deep hashing domain adaption deep hashing and cross-modal deep hashing are also included.}

Deep supervised hashing has been explored over a long period.
The design of the deep supervised hashing method mainly includes two parts: the design of the network structure and the design of the loss function. For small datasets like MINST \cite{lecun1998gradient} and CIFAR-10 \cite{krizhevsky2009learning}, shallow architecture such as AlexNet \cite{krizhevsky2012imagenet} and CNN-F \cite{chatfield2014return} are widely used. While for complex datasets like NUS-WIDE \cite{chua2009nus} and COCO \cite{lin2014microsoft}, deeper architecture such as VGG \cite{simonyan2014very} and ResNet50 \cite{he2016deep} are needed. The loss objectives are designed with the intention of maintaining similarity structures. These methods~\cite{li2021deep,cao2018deep} usually aim to narrow the difference between the similarity structures in the original and Hamming spaces. Researchers usually obtain the similarities in the original space by using label information in supervised scenarios, which is widely studied in different deep hashing methods. Hence how obtaining the similarities of learned hash codes are important for different algorithms. {We further categorize the deep supervised hashing algorithms according to how measuring the similarities of learned hash codes into four classes, i.e., pairwise methods, ranking-based methods, pointwise methods and quantization.} For each manner, we comprehensively analyze how the related articles design the optimization objective and take advantage of semantic labels, as well as what additional tricks are used.

\begin{figure*}[t]
    \centering
    \includegraphics[width=0.9\textwidth,keepaspectratio=true]{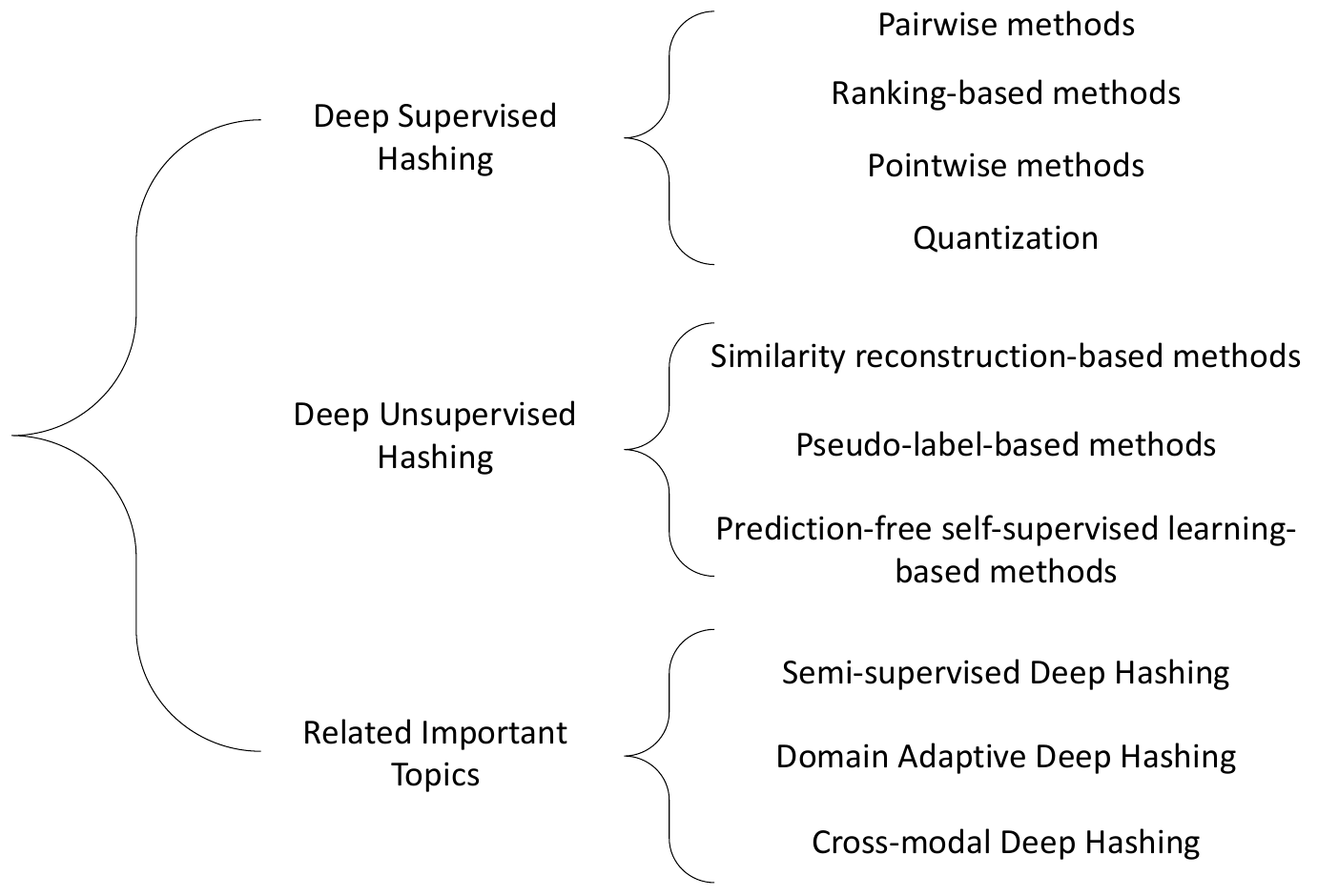}
    \caption{{The overall structure of this survey.}}\label{oveall}
\end{figure*}

Another area of research along this line is deep unsupervised hashing, which does not require any label information. Deep unsupervised hashing has drawn widespread attention recently
since it is easily applied in practice.
In unsupervised settings, the semantic information is usually derived from the relationship in the original space. Based on manners of learning semantic information, we categorize the deep unsupervised hashing algorithms into pseudo-label-based methods, similarity reconstruction-based methods and prediction-free self-supervised learning-based methods. 
In addition, we also introduce some other related important topics such as semi-supervised deep hashing, domain adaptation deep hashing and multi-modal deep hashing methods. The overall structure of this survey is shown in Fig. \ref{oveall}. 
Meanwhile, we also present some commonly used public datasets and the scheme to measure the performance of deep hashing algorithms. At last, a comparison of some key algorithms was given.

Compared to other surveys on hashing methods \cite{wang2014hashing,wang2015learning,wang2017survey,cao2017binary}, our paper mainly centers on recent deep hashing methods rather than traditional hashing methods and how they optimize the hashing network. Moreover, we study both deep supervised hashing as well as deep unsupervised hashing extensively. Finally, we classify two topics in a brand-new view based on the different manners of optimization.  
As far as we know, this is the most comprehensive survey about deep hashing, which is beneficial to researchers in understanding the mechanisms and trends of deep hashing.

\section{Background}
\subsection{Nearest Neighbor Search}
Given a $d$-dimensional Euclidean space $\mathbb{R}^{d}$, the nearest neighbor search aims to find the sample $\text{NN}(\bm{x}_r)$ in a finite set $\Pi \subset \mathbb{R}^{d}$ such that:
\begin{equation}
    \begin{aligned}
        \text{NN}(\bm{x}_r) = \arg\min_{\bm{x}_b \in \Pi} \rho(\bm{x}_r,\bm{x}_b),
    \end{aligned}
\end{equation}
in which $\bm{x}_r\in \mathbb{R}^{d}$ represents the query point. Note that $\rho$ could be any metrics such as Euclidean distance, cosine distance along with general $\ell_{p}$ distance. Many exact nearest neighbor search methods \cite{friedman1977algorithm} have been developed by the researchers, which works quite well when $d$ is small. However, nearest neighbor search is intrinsically costly due to the curse of dimensionality \cite{beyer1999nearest,bohm2001searching}. Although KD-tree can be extended to high-dimensional situations, its efficiency is far from satisfactory. 

To solve this problem, a series of algorithms for approximate nearest neighbors have been proposed \cite{datar2004locality,gionis1999similarity,muja2009fast,jegou2010product}. The principle of these methods is to find the nearest point with a high probability, rather than to find the nearest point accurately. These ANN algorithms are mainly divided into three categories: hashing-based methods \cite{datar2004locality,andoni2006near,lv2007multi}, product quantization based methods \cite{jegou2010product,ge2013optimized,kalantidis2014locally,zhang2014composite} and graph-based methods \cite{hajebi2011fast,malkov2014approximate,malkov2018efficient}. These algorithms have greatly improved the efficiency of searching while ensuring a relatively high accuracy, so they are widely used in the industry. Compared to the other two types of methods, hashing-based algorithms are the longest studied and the most studied by researchers because it has great potential in improving computing efficiency and reducing memory cost.

\subsection{Hashing Algorithms}
For nearest neighbor search, hashing algorithms are very efficient in terms of both computing and storage. 
Two main types of hashing-based search methods have been developed, i.e., hash table lookup~\cite{song2005fast,li2016deterministic} and hash code ranking~\cite{liu2016query,ji2014query}. 

The primary goal of hash table lookup is to decrease the number of distance calculations for speeding up searches. The structure of the hash table contains various buckets, each of which is indicated by one separate hash code.
Each point is associated with a hash bucket that shares the same hash code. Thus, the manner to learn hash codes for this kind of algorithm is to increase the likelihood of producing the same hash codes for adjacent points in the original space.
When a query is given, we can find the corresponding hash bucket according to the hash code of the query, so as to find the corresponding candidate set. After this step, we usually re-rank the points in the candidate set to get the final search target. However, the recall of selecting a single hash bucket as a candidate set will be relatively low.  Two methods are usually adopted to overcome this problem. The first method is to select some buckets that are close to the target bucket at the same time. The second method is to independently create multiple different hash tables according to different hash codes. Then we can select the corresponding target bucket from each hash table.

Hash code ranking is a relatively easier way than hash table lookup. When a query comes, we compute the Hamming distance between the query and each point in the searching dataset, then select the points with relatively smaller Hamming distances as the candidates for nearest neighbor search. After that, a re-ranking process by the original features is usually followed to obtain the final nearest neighbor. Different from hash table lookup methods, hash code ranking methods prefer hash codes that preserve the similarities or distances in the original space.

\subsection{Deep Neural Networks}
Deep neural networks~\cite{samek2016evaluating} have achieved significant success in various areas including computer vision~\cite{he2016deep,dosovitskiy2020image} and natural language processing~\cite{guan2019towards,strubell2019energy}. Early works such as deep belief network~\cite{hinton2006fast} and autoencoder~\cite{ng2011sparse} are mostly based on multi-layer perceptions. 
However, these networks do not show much better performance compared with traditional methods such as support vector machine and k-nearest neighbors algorithm. 
As convolutional neural networks have been introduced to process image data, various popular deep networks have been proposed and achieved promising results. AlexNet~\cite{krizhevsky2012imagenet} consists of five convolutional layers followed by three fully connected layers.
VGGNet~\cite{simonyan2014very} increases the model depth and improves the performance of image classification. NIN~\cite{networkinnetwork} is further proposed to promote the discriminability of image patches within the receptive field. Researchers have found that the depth of representations is the key to high performance for various visual recognition tasks. However, the problem of vanishing/exploding gradients makes it difficult to build very deep neural networks.
ResNet~\cite{he2016deep} tackles this problem by leveraging the residual learning to deepen the network and benefits from very deep models. Recently, Vision Transformer~\cite{dosovitskiy2020image} has achieved great success on image classification tasks due to its high model capacity and easy scalability. These powerful neural network architectures has become the backbone networks in various applications, including semantic segmentation \cite{strudel2021segmenter} and object detection \cite{detection}. In virtue of the strong representation ability of deep neural networks, deep hashing has shown great performance in image retrieval and drawn increasing attention recently. 

\begin{table}[t]
\centering
\caption{Summary of symbols and notation.\label{tab:sym}}
	\begin{tabular}{c|c}
		\hline
		Symbol & Description   \\
		\hline
	    $\bm{x}_i$ ($\bm{X}$) & input images (in matrix form) \\
	    $\bm{b}_i$ ($\bm{B}$) & output hash codes (in matrix form) \\
	    $\bm{h}_i$ ($\bm{H}$) & network outputs (in matrix form) \\
	    $\bm{y}_i$ ($\bm{Y}$) & one-hot image labels (in matrix form) \\
	     $\Psi(\cdot)$ &  hashing network \\
	    $N $ & the number of input images\\
	   
	     $L$ & hash code length \\
	     $\mathcal{E}$ & a set of pair items \\
	    $s_{ij}^o$ & the similarity of item pair $(\bm{x}_i, \bm{x}_j)$ in the input space \\
	    $s_{ij}^h$ & the similarity of item pair $(\bm{x}_i, \bm{x}_j)$ in the Hamming space \\
	    $d_{ij}^o$ & the distance of item pair $(\bm{x}_i, \bm{x}_j)$ in the input space \\
	    $d_{ij}^h$ & the distance of item pair $(\bm{x}_i, \bm{x}_j)$ in the Hamming space \\
	    $\epsilon $ & margin threshold parameter \\
	    $\bm{W}$ & weight parameter matrix \\
	    $\Theta$ & set of neural parameters \\
	   
		\hline
	
	\end{tabular}
\end{table}

\subsection{Learning to Hash}

Given an input item $\bm{x}$, learning to hash aims to obtain a hash function $f$, which maps $\bm{x}$ to a binary code $\bm{b}$ for the convenience of the nearest neighbor search. The hash codes obtained by a good hash function should preserve the distance order in the original space as much as possible, i.e., those items that are close to the specific query in Hamming space should also be close to the query in the original space. Many traditional hash functions include spherical function, linear projection and even a non-parametric function. A wide range of traditional hashing methods~\cite{wang2017survey,strecha2011ldahash,he2010scalable,weiss2009spectral,gionis1999similarity,gong2012iterative,liu2012supervised,shen2015supervised,zhang2021probability} have been proposed by researchers to learn compact hash codes, and achieved significant progress. {For instance, AQBC~\cite{gong2012angular} utilizes the angle between two vectors to measure the similarity and
maps feature vectors into the most similar vertices of a binary hypercube. FSDH~\cite{gui2017fast} regresses the semantic labels of samples to their binary codes and optimizes the hash codes in an alternative manner.
For a more comprehensive understanding, refer to a survey paper~\cite{wang2015learning}.}
However, these simple hash functions do not scale well for huge datasets. For the strong representation ability of deep learning, more and more researchers pay attention to deep supervised hashing and develop a range of promising methods. These methods generally achieve better performance than traditional methods.

\section{Deep Supervised Hashing}

{In this article, we first talk about deep supervised hashing methods, which are the basis of the subsequent deep unsupervised hashing techniques.}

\subsection{Overview}
\begin{figure*}[t]
    \centering
    \includegraphics[width=13cm,keepaspectratio=true]{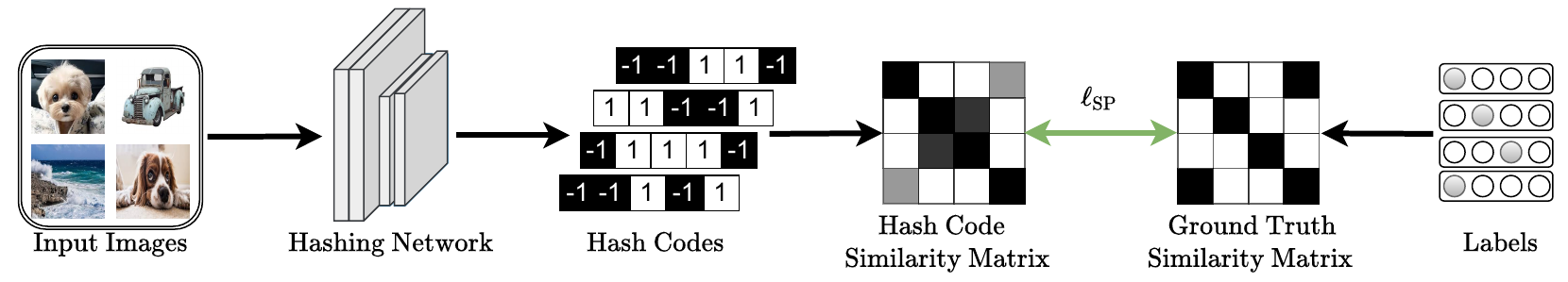}
    \caption{{Basic Framework of Deep Supervised Hashing with Pairwise Similarity Measurement. The hash codes are produced by a hashing network. Afterwards, the pairwise similarity information of hash codes and ground truth is matched and thus get similarity preserving loss. More details will be discussed in Sec. \ref{sec:pairwise}.}   }
    \label{fig:my_label}
\end{figure*}

Deep supervised hashing uses deep neural networks as hash functions, which can generate hash codes in an end-to-end manner. We focus on the following four key problems: (1) what deep neural network architecture is adopted; (2) how to design the loss function for preserving the similarity structure; (3) how to optimize the deep neural network with the discretization problem; (4) what other skills can be used to improve the performance. We first answer the first three problems in a nutshell and the last problem is left in the subsequent detailed introduction. Fig. \ref{fig:my_label} shows a representative framework of deep supervised hashing.

\subsubsection{Network Architecture}
Traditional hashing methods usually utilize linear projection and kernels, which {show} poor representation ability. After AlexNet and VGGNet \cite{krizhevsky2012imagenet,simonyan2014very} were proposed, deep learning shows its superiority in computer vision, especially for classification problems. And more and more experiments have proved that the deeper the network, the better the performance. As a result, ResNet \cite{he2016deep} takes advantage of residual learning, which can train very deep networks, achieved significantly better results. After that, ResNet and its variants have become basic architectures in deep learning \cite{he2016deep,huang2017densely,xie2017aggregated}. The latest researches often utilize the popular architectures with pre-trained weights in large datasets such as ImageNet, following the idea of transfer learning. Most of the researchers utilize shallower architectures such as AlexNet, CNN-F and design stacked convolutional neural networks for simple datasets, e.g., MNIST, CIFAR-10. Deeper architectures such as VGGNet and ResNet50 are often utilized for complex datasets such as NUS-WIDE and COCO. 
To be more precise, for deep supervised hashing methods, the hashing network is usually modified from these aforementioned standard networks by replacing the classification head with a fully-connected layer containing $L$ units for hash code learning. The network outputs are usually continuous codes.
The hash codes can be obtained using a sign activation. Graph neural networks, which capture the dependence between the nodes of graphs via message passing mechanisms, have been popular in various applications. They have also been adopted in recent hashing methods to learn the correlation of datasets~\cite{tan2020learning,wang2020learning,cui2021efficient}.

The architecture of the hashing network is one of the most important factors for deep supervised hashing, and it affects both the accuracy of the search and the time cost of inference. If the architecture degenerates into MLP or linear projections, deep supervised hashing will degrade into traditional hashing methods. Although the deeper the network architecture, the greater the search accuracy, it also increases the time cost. We think that the architecture needs to be considered combined with the complexity of datasets. As we know, the majority of existing deep hashing methods can use any network architecture as needed. Therefore, we do not adopt the network architectures for categorizing the deep supervised hashing algorithms. 

\subsubsection{Similarity Measurement and Objective Function}
We first provide formal notations and key concepts in Table \ref{tab:sym} for the sake of clarity. $\mathcal{X}=\{\bm{x}_i\}_{i=1}^{N}$ is denoted as the training set. $\mathcal{H}=\{\bm{h}_i\}_{i=1}^{N}$ denotes the outputs of the hashing network, i.e., $\bm{h}_i= \Psi(\bm{x}_i)$. $\mathcal{B}=\{\bm{b}_i\}_{i=1}^{N}$ is the obtained binary codes. We denote the similarity between pair of items $(\bm{x}_i,\bm{x}_j)$ in the input space and Hamming space as $s_{ij}^o$ and $s_{ij}^h$, respectively. 
In the input space, the similarity is the ground truth, which is mainly based on sample distance $d_{ij}^o$ and semantic labels. The former refers to the distance of features, e.g. Euclidean distance $||\bm{x}_i-\bm{x}_j||_2$, and the similarity can be computed using Gaussian function or Characteristic function, i.e., \(\exp \left(-\frac{\left(d_{i j}^{o}\right)^{2}}{2 \sigma^{2}}\right)\) and $I_{d_{ij}^o<\tau}$ where $\tau$ is a given threshold. The cosine similarity is also popular for measurement. The latter is more popular in deep supervised hashing, where the value is 1 if the two examples share a common semantic label and 0 vice visa. 

The pairwise distance $d_{ij}^h$ in the Hamming space is Hamming distance naturally, which is defined as follows: 
\begin{equation}
d_{i j}^{h}=\sum_{l=1}^{L} \delta\left[\bm{b}_{i}(l) \neq \bm{b}_{j}(l) \right],
\end{equation}
If the hash code is valued by 1 and 0, we have:
\begin{equation}
d_{i j}^{h}=\left\|\bm{b}_{i}-\bm{b}_{j}\right\|_{1},
\end{equation}
and it varies from $0$ to $L$. As a result, the similarity in this circumstance is denoted as $s_{ij}^h=(L-d_{ij}^h)/L$. If the code is valued by 1 and -1, we have:
\begin{equation}
d_{ij}^h=\frac{1}{2}(L-\bm{b}_i^T\bm{b}_j),
\end{equation}
The similarity is defined using the inner product, i.e., $
s_{i j}^{h}=(\bm{b}_{i}^{\top} \bm{b}_{j}+L)/2L
$. We can also extend this to the weighted circumstance. In formulation,
\begin{equation}
d_{i j}^{h}=\sum_{l=1}^{L} \lambda_{l} \delta\left[\bm{b}_{i}(l) \neq \bm{b}_{j}(l) \right],
\end{equation}
where each bit is associated a weight $\lambda_l$, and if the values of codes are 1 and -1, we have 
\begin{equation}
s_{i j}^{h}=(\bm{b}_{i}^{\top} \Lambda \bm{b}_{j}+ tr(\Lambda))/2tr(\Lambda),
\end{equation}
in which $\bm{\Lambda}=\text{diag}\left(\lambda_{1}, \lambda_{2}, \ldots, \lambda_{l}\right)$ is diagonal and $tr(\cdot)$ denotes the trace of the matrix. The weight of the associated hash bit fills each diagonal element of the matrix.

After defining the similarity measurement, we focus on the objective functions in deep supervised methods. A well-designed objective functions is one of the most important factors to promise the performance of deep supervised hashing. The main guideline for designing the objective function is to keep the similarity structure, which means to minimize the difference between the similarities in the original and Hamming spaces.
As a result, most of the objective functions contain the terms of similarity information. Among them, the typical loss functions are in a pairwise manner, making similar pairs of images have similar hash codes (small Hamming distance) and dissimilar pairs of images have dissimilar hash codes (large Hamming distance). 
Besides, a variety of researchers adopt ranking-based similarity preserving loss terms. For example, triplet loss is often used to maintain as much consistency as possible between the ordering of numerous items calculated from the original and Hamming spaces. There are also several listwise loss terms that consider the whole datasets for similarity preserving.

Besides similarity information, the pointwise label information is also well-explored in the design of the objective function. There are three popular ways to take advantage of label information summarized below. The first way is a regression on hash codes with labels. The label is encoded into one-hot format matrix and regression loss, i.e., $||\bm{Y}-\bm{W}\bm{H}||_F$ are added into the loss function. The second way is adding a classification layer after the hashing network, and a classification loss (e.g., cross-entropy loss) is added to the objective function. The last one is utilizing LabNet, which was first proposed in \cite{li2018deep}. LabNet aims to capture the ample semantic relationships among example pairs.

The quantization loss term is also commonly used in deep supervised hashing, especially in quantization-based hashing methods. The typical form of quantization is to penalize the distance between continuous codes (i.e., network outputs) and binary codes. As a common technique in deep hashing, {bit balance} loss penalizes the situation that each bit has a large chance of being $1$ or $-1$ among the whole dataset. Several regularization losses can be added to the loss function, which is also important for improving the performance.

\subsubsection{Optimization Algorithm}
It is difficult to optimize the hashing network parameters because of the vanishing gradient issue resulting from the sign activation function which is used to obtain binary hash codes. Specifically, the sign function is in-differentiable at zero and its gradient is zero for all nonzero input, which is fatal to the hashing network using gradient descent for optimization.

Almost all the works adopt that continuous relaxation by smoothing the sign function using the sigmoid function or the hyperbolic tangent function, and apply sign function to obtain final binary codes later in the evaluation phase. The first typical way is quantization function by adding a penalty term in loss function, which is often formulated as $|||\bm{h}_i|-\bm{1}||_1$, or $-||\bm{h}_{i}||$ with tanh activation. This penalty term helps the neural network to obtain $sgn(\bm{h_i})\approx \bm{h_i}$. {Note that} this loss can be considered as a prior for every binary code $h_i$ on basis of a variant of certain distribution, e.g., bimodal Laplacian and Cauchy distribution. From this view, we can get a few variants, e.g., pairwise quantization \cite{zhu2016deep} and Cauchy quantization loss \cite{cao2018deep}. If the loss function is a non-smooth function and its derivative is hard to calculate, a modified version can be adopted instead, e.g., \(|x| \approx \log (\cosh x)\)~\cite{zhu2016deep}. The second way is an alternative scheme, which resolves the optimization into several sub-problems. Then, these sub-problems could be iteratively settled through alternating the minimization of objectives. In this alternative process, backpropagation can only work in one sub-problem and the other sub-problems can be solved by other optimization methods. For example, DSDH \cite{li2017deep} utilizes the discrete cyclic coordinate descend algorithm. These methods can keep the discrete constraint during the whole optimization process while it can not lead to end-to-end training, which has limited application for solving the unknown sub-problems. The third method is named continuation which utilizes a smoothed function $y= tanh(\beta x)$ to approach the discrete activation function by increasing $\beta$ \cite{cao2017hashnet}. There are some other ways to solve this problem by changing the calculation and the propagation of gradients, e.g., Greedy Hash \cite{su2018greedy} and Gradient Attention Network \cite{huang2019accelerate}, which improve the effectiveness and accuracy of deep supervised hashing. 

\subsubsection{Summarization}
In this survey, we divide the current methods into the following four classes mainly based on how to measure the similarities in the Hamming space: the pairwise methods, the ranking-based methods, the pointwise methods and the quantization methods.
The quantization methods are separated from the pairwise methods due to their specificity. The key motivation we select how to measure the similarities of the learned hash codes for categorization is that the fundamental core of learning to hash is to maintain the similarity structure and the manners of similarity measurement decide the loss functions in deep supervised hashing. Additionally, neural network architectures, optimization manners as well as other skills are also significant for the retrieval performance. For each class, we will discuss the corresponding deep hashing methods in detail one by one. The detailed summarization of these methods is shown in Table \ref{table_example}.

\begin{table}

\renewcommand{\arraystretch}{1.3}
\tiny
\caption{A Summary of Deep Supervised Hashing Methods w.r.t the Different Manner of Similarity Measurement (Pairwise Methods, Ranking-based Methods and Pointwise Methods), Binarization as well as Other Skills. Drop = Drop the sign operator in the neural network and treat the binary code as an approximation of the network output, Two-step = Two-step optimization. {Reg. = Regression, Quan. = Quantization Loss, Cla. = Classification, Ind. = Independence, Regu. = Regularization, Bal. = Balance, Diff. = Difference Loss, Prod. = Product Loss, Like. = Likelihood Loss.  } }\label{table_example}
\centering
\begin{tabular}{l|c|c|c|c|c}
\hline
 Approach  & Pairwise & Ranking-based & Pointwise  & Binarization & Other skills\\
\hline
            SDH \cite{erin2015deep}      & Prod. & - & - & Quan. & Bit Bal. + Orthogonality   \\

            DSH \cite{liu2016deep}  & Prod. + Margin & - &  - & Quan.& - \\
            PCDH \cite{chen2019deep1}      & Prod. + Margin & - & Cla. Layer & Drop & Pairwise Correlation \\
            WMRSH \cite{li2019weighted}      & Prod. + Margin & - & Cla. Layer & Quan. & Bit and Table Weight \\
            
            SHBDNN \cite{do2016learning}      & Diff. & - & - & Quan. + Alternation & Bit Bal. + Independence \\
            DDSH \cite{jiang2018deep}      &Diff.  & - & -  & Alternation  & Splitting Training Set \\
            CNNH \cite{xia2014supervised}   & Diff. & - & Part of Hash Codes & - & Two-step \\
            ADSH \cite{jiang2018asymmetric}      & Diff. & - & - & Quan. +Alternation & Asymmetry \\
            DIH \cite{wu2019deep}      &  Diff. & - & - & Quan.+Alternation & Incremental Part + Bit Bal.\\
            HBMP \cite{cakir2018hashing}      & Diff. & - & - & Drop & Bit Weight + Two-step \\
            DOH \cite{jin2018deep}         & Diff. & - & - & Ranking & FCN \\
            DPSH \cite{li2016feature}      & Like.  & - & - & Quan. & -\\
            DHN \cite{zhu2016deep}   & Like.  & - & - & Quan. + Smooth & - \\
            HashNet \cite{cao2017hashnet}      & Weighted Like.  & - & - & Tanh + Continuation & - \\
            DSDH \cite{li2017deep}   & Like.  & - & Linear Reg.+ L2 & Quan. + Alternation & - \\
            DAPH \cite{shen2017deep}      & Like.  & - & - & Quan. + Alternation & Bit Bal. + Ind.  \\

            DAgH \cite{yang2015supervised}   & Like. + Diff. & -  & - & - & Two-step  \\
            
            DCH \cite{cao2018deep}       &  Cauchy Like. & -  & - & Cauchy Quan. & - \\
           
            DJSEH \cite{li2018deep}      & Like. & -  & LabNet + Linear Reg. & Quan. & Two-step + Asymmetry \\
            ADSQ \cite{yang2019asymmetric}      & Diff.+ Like.  & - & LabNet & Quan.+Alternation & Bit Bal. + Two-step \\
            MMHH \cite{kang2019maximum}      & t-Distribution. Like. & - & - & Quan. & Semi-Batch Optimization \\
            DAGH \cite{chen2019deep}      & Like. & -  & Linear Reg. & Drop + Alternation & Reg. with Anchor Graph \\
            HashGAN \cite{cao2018hashgan}      & Weighted Like. & - & - & Cosine Quan. & GAN  \\
            DPH \cite{cao2018deep1}   & Priority Like. & -  & - & Priority Quan. & Priority CE Loss \\ 
            DFH \cite{li2019push}      & Like. + Margin & - & - & Quan. + Alternation & Quantized Center Loss \\  
            DRSCH \cite{zhang2015bit}      & Diff. + Margin & Triplet + Margin  & - & Drop & Bit Weight \\
            DNNH \cite{lai2015simultaneous}       & -  & Triplet + Margin & - & Piecewise Thresholding & - \\
            DSRH \cite{zhao2015deep}      & -  & Weighted Triplet & - & Quan. & Bit Bal. \\
            DTSH \cite{wang2016deep}      & - & Triplet + Like. & - & Quan. & - \\
            DSHGAN \cite{qiu2017deep}       & - & Triplet + Margin &  Cla. Layer & Drop & GAN \\

            AnDSH \cite{zhou2019angular}      & -  & Matrix Optimization & Angular-softmax & Drop & Bit Bal. \\
            HashMI \cite{cakir2019hashing}      & -  & Mutual Information & - & Drop & - \\
            TALR \cite{he2018hashing}      &  -& Relaxed AP + NDCG  & - & Tie-Awareness \\

            MLRDH \cite{liu2019mutual}      & - & - & Multi-linear Reg. & Alternation & Hash Boosting \\
            HCBDH \cite{chen2019hadamard}  & -     & - &  Cla. Layer & - & Hadamad Loss  \\
            DBH \cite{lin2015deep}      & - & - &  Cla. Layer & - & Transform Learning \\
            SSDpH \cite{yang2015supervised}    & - & - &  Cla. Layer & Quan. & Bit Bal. \\          
            VDSH \cite{zhang2016efficient}      & -& -  & Linear Reg. & Drop + Alternation & - \\
            PMLR~\cite{shen2019embarrassingly} & - & - & Cla. Layer & - & Distribution Regu. \\
            CSQ \cite{yuan2019central} & - & - & Center + Binary CE  & Quan. & - \\
            DPH \cite{fan}& -& -  & Center + Polarization  & - & - \\
            OrthHash \cite{hoe2021one}& - & - & Center + CE  & - & - \\
            PSLDH~\cite{tu2021partial} & - & - & Center + Partial Loss  & Quan. & - \\

            DVsQ \cite{cao2017deep}      & - & Triplet Loss+Margin & - &  Inner-Product Quan.&   Label Embeddings\\      
            DPQ \cite{klein2019end}      & - & - &  Cla. Layer & - & Joint Central Loss \\
            DSQ \cite{eghbali2019deep}      & - & - &  Cla. Layer  & Quan. + Alternation & Joint Central Loss \\
            SPDAQ \cite{chen2019similarity}      & Diff. & - &  Cla. Layer & Drop + Alternation & Asymmetry \\
            DQN \cite{cao2016deep}      &Diff.  & - & - & Product Quan. & Asymmetric Quan. Distance \\
            DTQ \cite{liu2018deep}      & - & Triplet + Margin & - & Weak-Orthogonal Quan. & Group Hard \\
         
            
          
\hline
\end{tabular}
\end{table}

\subsection{Pairwise Methods}\label{sec:pairwise}

We further divide the techniques which match the distances or similarities of image pairs derived from two spaces, i.e., the original space and the Hamming space into two parts as follows:

\begin{itemize}
	\item Difference loss minimization. {The kind of losses minimizes the difference between the similarities, i.e., \(\min \sum_{(i, j) \in \mathcal{E}}\left(s_{i j}^{o}-s_{i j}^{h}\right)^{2}\)~\cite{do2016learning,jiang2018deep,xia2014supervised,jiang2018asymmetric,wu2019deep,cakir2018hashing,jin2018deep}. $s_{i j}^{h}$ can be derived with inner product of binary codes, i.e., $s_{i j}^{h} = \bm{b}_i^T\bm{b}_j/L$ and $s_{i j}^{o}$ is now valued by $1$ or $-1$.
	However, binary optimization is difficult to implement. Early methods utilizes the relaxed outputs of neural networks to replace the hash codes, i.e., $s_{i j}^{h} = \bm{h}_i^T\bm{h}_j/L$~\cite{do2016learning,xia2014supervised}. Subsequent methods utilize a asymmetric manner to calculate the similarity, i.e., $s_{i j}^{h} = \bm{b}_i^T\bm{h}_j/L$, which releases the impact of quantization error~\cite{jiang2018deep,wu2019deep,jin2018deep}. There are also works combining both symmetric and asymmetric similarities~\cite{jiang2018asymmetric}. 
	Weighted bits are also introduced for adaptive similarity calculation~\cite{cakir2018hashing}. 
	Note that the difference losses can be transformed into a product form. Hence, we also categorize the methods minimizing product loss as this group. They usually adopt a loss in the product form, i.e., \(\min \sum_{(i, j) \in \mathcal{E}} s_{i j}^{o} d_{i j}^{h}\)~\cite{erin2015deep}, which expects that if the similarities in the original space are higher, the distances in the Hamming space should be less. Subsequent methods usually involve a margin in the loss for better relaxation~\cite{liu2016deep,chen2019deep1,li2019weighted}. 
	}

	\item Likelihood loss minimization. This kind of losses is derived from the probabilistic model. Given similarity matrix $\bm{S}=\{s_{ij}^o\}_{(i,j)\in \mathcal{E}}$ and hash codes $\bm{B} = [\bm{b}_1, \dots, \bm{b}_N]^T$, the posterior estimation of binary codes is formulated as follows:
	
\begin{equation}
    \begin{aligned}
    p(\bm{B}| \bm{S}) \propto p(\bm{S} |\bm{B}) p(\bm{B})=\prod_{(i, j) \in \mathcal{E}} p\left(s_{i j}^o | \bm{B}\right) p(\bm{B}),
    \end{aligned}
\end{equation}
where $p(\bm{B})$ denotes a prior distribution and $p(\bm{S}|\bm{B})$ is the likelihood. The conditional probability of $s_{ij}^o$ given their hash codes is denoted by $ p\left(s_{i j}^o |\bm{B}\right)$. Note that the $s_{ij}^h $ is derived from $\bm{B}$. In formulation, 
\begin{equation}\label{eq:3}
p\left(s_{i j}^{o} | \bm{B}\right)=p\left(s_{i j}^{o} | s_{i j}^{h}\right)=\left\{\begin{array}{cc}\sigma\left(s_{i j}^{h}\right), & s_{i j}^{o}=1 \\ 1-\sigma\left(s_{i j}^{h}\right), & s_{i j}^{o}=0\end{array}\right.
\end{equation}
in which $\sigma(x)=1/(1+e^x)$. From Eq. \ref{eq:3}, the probabilistic model expects the similarities in the Hamming space to be larger if the similarities in the original space are larger. The loss function is the negative log-likelihood (NLL)~\cite{li2016feature,zhu2016deep}, i.e.,
    
\begin{equation}
    \begin{aligned}
    \mathcal{L}_{NLL} = -\log p(\bm{S}|\bm{H})=\sum_{(i, j) \in \mathcal{E}}\log(1+e^  {s_{ij}^h})-s_{i j}^o s_{ij}^h.
    \end{aligned}
\end{equation}
{Similarly, the hashing network usually cannot directly obtain the hash codes. Hence these codes $\bm{B}$ will be replaced by the network outputs $\bm{H}$ to generate $s_{ij}^h$. The majority of methods adopt the symmetric similarities while several methods utilize the asymmetric form for similarity calculation~\cite{shen2017deep,li2018deep}.
However, the sigmoid function in Eq. \ref{eq:3} is not optimal and there are a number of works that utilize different tools to design valid probability functions, e.g., priority weighting~\cite{cao2018deep1}, Cauchy distribution~\cite{cao2017deep}, imbalance learning~\cite{cao2017hashnet} and t-Distribution~\cite{kang2019maximum}. Subsequent works combine label information with pairwise similarity learning for better semantic preserving~\cite{li2018deep,li2017deep,chen2019deep}. Li et al.~\cite{li2019push} associate the likelihood loss with Fisher's Linear discriminant, and introduce a margin for discriminative hash codes. Chen et al.~\cite{chen2019deep} reduce the computational cost by introducing anchors for similarity calculation. There are also some works combining both difference loss minimization and likelihood loss minimizing for comprehensive optimization~\cite{yang2019deep}.
} 

\end{itemize}

Although these methods in each group could utilize the same pairwise loss term, they may involve different architectures, optimization manners and regularization terms, as shown in Table \ref{table_example}. These details of different variants will be shown below.

\subsubsection{Difference Loss Minimization}
\emph{Deep Supervised Hashing} (DSH)~\cite{liu2016deep}. DSH unitizes a network consisting of three convolutional-pooling layers and two fully connected layers. {Recall that the outputs of the hashing network are $\{\bm{h}_i \}_{i=1}^N$.} The origin pairwise loss function is defined as:

\begin{equation}
    \begin{aligned}
 \quad & \mathcal{L}_{\text{DSH}}=\sum_{(i, j) \in \mathcal{E}}\frac{1}{2}s_{ij}^od_{ij}^h+\frac{1}{2}(1-s_{ij}^o)[\epsilon-d_{ij}^h]_+\\
 & \mbox{s.t.} \forall \quad  \bm{h}_i,\bm{h}_j \in \{-1,1\}^L\\
    \end{aligned}
\end{equation}
where $d_{ij}^h = ||\bm{h}_i-\bm{h}_j||_2^2$, $[\cdot]_+$ denotes $\max(\cdot,0)$ and $\epsilon>0$ is a given threshold parameter. 
The loss function obeys a distance-similarity product minimization formulation that expects similar examples mapped to similar binary codes and rewards dissimilar examples transferred to distinct binary codes when the Hamming distances are smaller compared with the margin threshold $m$. It is noticed that when $d_{ij}^h$ is larger than $m$, the loss does not produce gradients. This idea is similar to the hinge loss function. 

As we discuss before, DSH relaxes the binary constraints and a regularizer is added to the continuous outputs of the hashing network, which approximates the binary codes, i.e., $h\approx sgn(h)$. The pairwise loss is rewritten as:

\begin{equation}
    \mathcal{L}_{\text{DSH}}=\frac{1}{2}s_{ij}^o||\bm{h}_i-\bm{h}_j||_2^2+\frac{1}{2}(1-s_{ij}^o)[\epsilon-||\bm{h}_i-\bm{h}_j||_2^2]_+ +\lambda_1\sum_{k=i,j}|||\bm{h}_k|-\bm{1}||_1,
\end{equation}
where $\bm{1}$ denotes a all-one vector and $||\cdot||_p$ produces the $\ell_{p}$-norm of the vector. $\lambda_1$ is a parameter to balance the effects of the regularization loss. DSH does not utilize saturating non-linearities because it may slow down the training process. With the above loss function, the neural network is able to be trained with an end-to-end back propagation algorithm. For the {evaluation} process, the binary codes can be derived using the sign activation function. 
DSH is a straight-forward deep supervised hashing method in the early period, and its idea originates from Spectral Hashing \cite{weiss2009spectral} but with a deep learning framework.  

\emph{Pairwise Correlation Discrete Hashing} (PCDH) \cite{chen2019deep1}. PCDH utilizes four fully connected layers after the convolutional-pooling layer, named deep feature layer, hash-like layer, discrete hash layer as well as classification layer, respectively. The third layer can directly generate discrete hash code. {Different from} DSH, PCDH leverages $\ell_2$ norm of deep features and hash-like codes. Besides, the classification loss is included in the final function:

\begin{equation}
\begin{aligned} 
\mathcal{L}_{\text{PCDH}}&= \mathcal{L}_{s}+\lambda_1 \mathcal{L}_{p}+\beta \mathcal{L}_{l} \\
&=\sum_{(i, j) \in \mathcal{E}}\left(\frac{1}{2}(1-s_{i j}^o)[\epsilon-\|\bm{h}_{i}-\bm{h}_{j}\|_{2}^{2}]_+^{2}+\frac{1}{2} s_{i j}\|\bm{h}_{i}-\bm{h}_{j}\|_{2}^{2}\right) \\ 
&+\lambda_1\sum_{(i, j)\in \mathcal{E}}\left(\frac{1}{2}\left(1-s_{i j}^o\right)[\epsilon-||\bm{z}_i-\bm{z}_j||_{2}^{2}]_+^{2}+\frac{1}{2} s_{i j}^o\|\bm{z}_i-\bm{z}_j\|_{2}^{2}\right)\\
&+\lambda_2\left(\sum_{i=1}^{N} \phi(\bm{w}_{i}^{T} \bm{b}_{i}, \bm{y}_{i})+\sum_{j=1}^{N} \phi(\bm{w}_{j}^{T} \bm{b}_{j}, \bm{y}_{j})\right)
\end{aligned}
\end{equation}
{where $\bm{z}_i,\bm{h}_i$ and $\bm{b}_i$ denote the outputs of the first three fully connected layers. The last term is the classification cross-entropy loss\footnote{In our survey, $\{ \lambda_1, \lambda_2, \lambda_3, \cdots \} $ always denote the balance coefficients. }. Note that the second term is called pairwise correlation loss, which guides the similarity learning of deep features to avoid overfitting.} The classification loss provides semantic supervision, which helps the model achieve competitive performance. Besides, PCDH proposes a pairwise construction module named Pairwise Hard, which samples positive pairs with the maximum distance between deep features and negative pairs with the distances smaller than the threshold randomly. It is evident that Pairwise Hard chooses the hard pairs with the large loss for effective hash code learning. 

\emph{Supervised Deep Hashing} (SDH) \cite{erin2015deep}. SDH utilizes the fully-connected neural network for deep hashing and has a similar loss function except for a term that enforces a relaxed orthogonality constraint on all projection matrices (i.e., weight matrices in a neural network) for the property of fully-connected layers. Bit balance regularization is also included which will be introduced in Eq. \ref{eq:8}. 

\emph{Supervised Hashing with Binary Deep Neural Network} (SH-BDNN)~\cite{do2016learning}. The architecture of SH-BDNN is stacked by a fully connected layer, in which $\bm{W}_i$ denotes the weights in the i-th layer. {SH-BDNN not only considers the bit balance, i.e., each bit obeys a uniform distribution, but also considers the independence of different hash bits.} Given the hash code matrix $\bm{B} = [\bm{b}_1, \dots, \bm{b}_N]^T$, the two conditions are formulated as 
\begin{equation}
    \bm{B}^T\bm{1}=\bm{0}, \frac{1}{N}\bm{B}^T\bm{B}=\bm{I}
\end{equation}
where $\bm{1}$ is a $L$-dimension vector whose elements are all one, and $\bm{I}$ is an identity matrix of size $N$ by $N$. The loss function is 

\begin{equation}\label{eq:8}
\begin{aligned} 
 \quad & \mathcal{L}_{\text{SH-BDNN}}=\frac{1}{2 N}||\frac{1}{L}\bm{H}\bm{H}^{T} -\bm{S}||^{2}\\
&\quad \quad \quad \quad +\frac{\lambda_{1}}{2}
\sum_{k=1}^{K-1}||\bm{W}^{(k)}||^{2} +\frac{\lambda_{2}}{2N}||\bm{H}-\bm{B}||^{2} \\
&\quad \quad \quad \quad +\frac{\lambda_{3}}{2}||\frac{1}{N} \bm{H}^{T}\bm{H}-\bm{I}||^{2} +\frac{\lambda_{4}}{2 N}||\bm{H}^T \bm{1}||^{2} \\ 
 & \mbox{s.t.} \quad \bm{B} \in\{-1,1\}^{N \times L} 
\end{aligned}
\end{equation}
{$\bm{H}$ is stacked by the outputs of network and $\bm{B}$ is stacked by the binary codes} to be optimized from the Equation \ref{eq:8}. $\bm{S}$ is the pairwise similarity matrix valued 1 or -1. The first term is similarity difference loss minimization, the second term is the $\ell_{2}$ regularization, the third term is the quantization loss, the last two terms are to punish the correlation and the imbalance of bits respectively. Note that the $\bm{B}$ is not the sign of $\bm{H}$. As a result, the loss function is optimized by updating the network parameter and $\bm{B}$ alternatively. To be specific, $\bm{B}$ is optimized with a fixed neural network while the neural network is trained with fixed $\bm{B}$ alternatively. SH-BDNN has a well-designed loss function which follows Kernel-based Supervised Hashing \cite{liu2012supervised}. However, the architecture {does not} include the popular convolutional neural network and it is not an end-to-end model. {As a result, the efficiency of this model is low in large-scale datasets.} 

\emph{Convolutional Neural Network Hashing} (CNNH) \cite{xia2014supervised}. CNNH is the earliest deep supervised hashing framework to our knowledge. It {adopts} a two-step strategy. In the first step, it optimizes the objective function using a coordinate descent strategy as follows:
\begin{equation}
 \mathcal{L}_{\text{CNNH}}=||\frac{1}{L}\bm{H}\bm{H}^{T} -\bm{S}||^{2}
\end{equation}
which generates approximate binary codes. In the second step, CNNH utilizes obtained hash codes to train the convolutional neural network with $L$ output units. Besides, if class labels are available, the fully-connected layer with $K$ output units is added, which correspond to the $K$ class labels of images and the classification loss is added to the loss function. Although CNNH {uses labels in a clumsy manner}, this two-step strategy is still popular in deep supervised hashing and inspires many other state-of-the-art methods.

\emph{Deep Discrete Supervised Hashing} (DDSH) \cite{jiang2018deep}. DDSH uses a column-sampling manner for partitioning the training data into $\{\bm{x}_i\}_{i\in\Omega}$ and $\{\bm{x}_i\}_{i\in\Gamma}$, where $\Omega$ and $\Gamma$ are the indexes. The loss function is designed {in} an asymmetric form:
{
\begin{equation}
\mathcal{L}_{\text{DDSH}}= \sum_{i\in\Omega,j\in\Gamma}\mathcal{L}(s_{ij}^o - \bm{b}_i^T \bm{h}_j)^2 +\sum_{i,j\in \Omega}\mathcal{L}(s_{ij}^o - \bm{b}_i^T \bm{b}_j)^2 
\end{equation}
}
where $\bm{b}_i$ and $\bm{h}_i$ are the binary code to be optimized and the output of the network, respectively. $\bm{b}_i$ and $\bm{h}_i$ are updated alternatively {following} \cite{do2016learning}. It is notable because DDSH takes an asymmetric strategy for learning to hash, which aids in both binary code generation and continuous feature learning through the pairwise similarity structure.

\emph{Hashing with Binary Matrix Pursuit} (HBMP) \cite{cakir2018hashing}. HBMP also takes advantage of the two-step strategy introduced above. {Different from CNNH}, HBMP utilizes the weighted Hamming distances and adopts a different traditional hashing algorithm called binary code inference to get hash codes. In the first step, the objective function is written in the following equation:
\begin{equation}
\mathcal{L}_{\text{HBMP}}=\frac{1}{4}\sum_{i,j}\left(\bm{b}_i^T\Lambda \bm{b}_j -s_{ij}^o\right)^2,
\end{equation}
where $\bm{\Lambda}$ is a diagonal weight matrix. It is noticed that the similarity matrix {with each element} $S^h_{ij}=\bm{b}_i^T\Lambda \bm{b}_j$ can be approximated by a step-wise algorithm. HBMP also trains a convolutional neural network by the obtained hash codes with point-wise hinge loss and shows that deep neural networks help to simplify {the optimization problem} and get robust hash codes.

\emph{Asymmetric Deep Supervised Hashing} (ADSH) \cite{jiang2018asymmetric}. ADSH considers the samples in the database and query set using an asymmetric manner, which can help to train the model more effectively, especially for large-scale nearest neighbor search. ADSH contains two critical components, {i.e., a} feature learning part and a loss function part. The first one is to utilize a hashing network to learn discrete codes for queries. The second one is used to directly learn discrete codes for database points by minimizing the same objective function with supervised information. The loss function is formulated as: 

\begin{equation}
    \begin{aligned}
 & \mathcal{L}_{\text{ADSH}}=\sum_{i \in\Omega, j\in \Gamma} \left(\bm{h}_i^T\bm{b}_j-L s_{ij}^o\right)^2, \\
 & \mbox{s.t.}\quad \bm{b}_j\in \{-1,1\}^L,
    \end{aligned}
\end{equation}
where $\Omega$ is the index of query points, $\Gamma$ is the index of database points.
Network parameters $\Theta$ and binary codes $\bm{b}_j$ are updated alternatively following SH-BDNN \cite{do2016learning} during the optimization process. 
If only the database points are available, we let $\Omega\subset \Gamma$ and add a quantization loss $\sum_{i \in \Omega}\left(\bm{b}_i-\bm{h}_i\right)^2$ with the coefficient $\gamma$. {This asymmetric strategy combines deep hashing and traditional hashing, which can help achieve better performance.} 

\emph{Deep Incremental Hashing Network} (DIHN) \cite{wu2019deep}. DIHN tries to learn hash codes in an incremental manner. Similar to ADSH \cite{jiang2018asymmetric}, the dataset {is} divided into two parts, {i.e.,} original and incremental databases respectively. When a new image comes from an incremental database, its hash code is learned while keeping the hash codes of the original database unchanged. The optimization process still uses the strategy of alternately updating parameters. 

\emph{Deep Ordinal Hashing} (DOH) \cite{jin2018deep}. DOH generates ordinal hash codes by taking advantage of both local and global features. Specifically, two subnetworks learn the local semantics using a spatial attention module-enhanced fully convolutional network and the global semantics using a convolutional neural network, respectively. Afterward, the two outputs are combined to produce $R$ ordinal outputs $\{\bm{h_i}^r\}_{r=1}^R$. For each segment $\bm{h_i}^r$, the corresponding hash code can be obtained as follows:
\begin{equation}
    \begin{aligned}
        \bm{b}_i^r &= \arg\max_{\bm{\theta}} \bm{\theta}^T\bm{h}_i,\\
        \text{s.t.}& \bm{\theta}\in\{0,1\}^L, \|\bm{\theta}\|_{1}=1,  
   \end{aligned}
\end{equation}
The full hash code can be obtained by concatenating {$\{\bm{b}_i^r\}_{r=1}^R$}. DOH adopts an end-to-end ranking-to-hashing framework, which {avoids} using the undifferentiable sign function. {Furthermore}, it uses a relatively complex network that is {able} to handle large datasets with higher performance.  

\subsubsection{Likelihood loss minimization}
\emph{Deep Pairwise Supervised Hashing} (DPSH) \cite{li2016feature}. 
DPSH adopts CNN-F \cite{chatfield2014return} as the {backbone of the hashing network} and the standard form of likelihood loss based on similarity information. Besides similarity information, quantization loss is also introduced to the final loss function, i.e.,
\begin{equation}
 \mathcal{L}_{DPSH}=-\sum_{(i,j)\in\mathcal{E}}\left(s_{i j}^o s_{ij}^h-\log \left(1+e^{s_{ij}^h}\right)\right) +\lambda_1 \sum_{i=1}^{N}||\bm{h}_{i}-sgn(\bm{h}_{i})||_{2}^{2},
\end{equation}
where $s_{ij}^h=\frac{1}{2}\bm{h}_{i}^T\bm{h}_{j}$ and $\bm{h}_{i}$ is the output of the {hashing} network. Although triplet loss was popular at that time, DPSH {adopts the pairwise form to simultaneously learn deep features} and hash codes, which improves both accuracy and efficiency. This likelihood loss function can easily introduce different Bayesian priors, making it flexible in {applications and achieving better performance than different loss functions. }

\emph{Deep Hashing Network} (DHN) \cite{zhu2016deep}. It has a {similar likelihood} loss function to DPSH. {Differently}, DHN {considers} the quantization loss as Bayesian prior and proposes a bimodal Laplacian prior for the output $\bm{h}_i$, i.e.,
\begin{equation}
    \begin{aligned}
p\left(\bm{h}_{i}\right)=\frac{1}{2 \epsilon} \exp \left(-\frac{\left\|\left|\bm{h}_{i}\right|-\bm{1}\right\|_{1}}{\epsilon}\right),
    \end{aligned}
\end{equation}
and {the} negative log likelihood (i.e. quantization loss) is 

\begin{equation}
    \begin{aligned}
\mathcal{L}_{Quan}=\sum_{i=1}^N |||\bm{h}_i-\bm{1}||_1,
    \end{aligned}
\end{equation}
which can be smoothed by a smooth surrogate \cite{hyvarinen2009natural} into 

\begin{equation}
    \begin{aligned}
\mathcal{L}_{Quan}=\sum_{i=1}^N \sum_{l=1}^L log (cosh(|h_{il}|-1)),
    \end{aligned}
\end{equation} 
where $\bm{h}_{ik}$ is the k-th element of $\bm{h}_i$. We notice that the DHN replaced $\ell_2$ norm (ITQ quantization error \cite{gong2012iterative}) by $\ell_1$ norm. \cite{zhu2016deep} also shows that the $\ell_{1}$ norm is an upper bound of the $\ell_{2}$ norm, and the $\ell_{1}$ norm encourages sparsity and is easier to optimize. 

\emph{HashNet} \cite{cao2017hashnet}. As a variant of DHN, HashNet {considers} the imbalance training problem that the positive pairs are much more than the negative pairs. {Hence,} it adopts Weighted Maximum LikeLihood (WML) loss with different weights for each image pair. The weight is formulated as 
\begin{equation}
    \begin{aligned}
w_{i j}=c_{i j} \cdot\left\{\begin{array}{ll}{|\mathcal{S}| /\left|\mathcal{S}_{1}\right|,} & {s_{i j}^o=1} \\ {|\mathcal{S}| /\left|\mathcal{S}_{0}\right|,} & {s_{i j}^o=0}\end{array}\right.
    \end{aligned}
\end{equation}
where \(\mathcal{S}_{1}=\left\{(i,j)\in\mathcal{E}: s_{i j}^o=1\right\}\)
comprises similar image pairs while $\mathcal{S}_0 = \mathcal{E}/\mathcal{S}_1 $ comprises dissimilar image pairs. \(c_{i j}=\frac{\bm{y}_{i} \cap \bm{y}_{j}}{\bm{y}_{i} \cup \bm{y}_{j}}\) for multi-label datasets and equals 1 for single-label datasets. Besides, the sigmoid function in condition probability is substituted by $1/1+e^{-\alpha x}$ called adaptive sigmoid function which equals to adding a hyper-parameter into the hash {code} similarity computation, i.e., $s_{ij}^h=\alpha \bm{b}_i^T\bm{b}_j$.
Different from other methods, HashNet continuously approximates sign function through the {hyperbolic tangent function}:
\begin{equation}
\lim _{\beta \rightarrow \infty} \tanh (\beta z)=\operatorname{sgn}(z).
\end{equation}
The activation function for outputs is $\tanh(\beta_t \cdot)$ through updating $\beta_t\rightarrow \infty$ step-wise and the optimal network with $\operatorname{sgn}(\cdot)$ can be derived. Besides, this operation can be illustrated using multi-stage pretraining, which means that the deep network using activation function $tan(\beta_{t+1}\cdot)$ is initialized using the well-trained network using activation function $tan(\beta_t\cdot)$. The two skills proposed by HashNet greatly increase the performance of deep supervised hashing. 

\emph{Deep Priority Hashing} (DPH) \cite{cao2018deep1}. {DPH} also adds different weights to different image pairs, but reduces the {weights} of pairs with higher confidence, which is similar to AdaBoost~\cite{schapire2013explaining}. The difficulty is measured by $q_{ij}$, which indicates how difficult a pair is classified as similar when $s_{ij}^o=1$ or classified as dissimilar when $s_{ij}^o=0$. In formulation, 

\begin{equation}
\begin{aligned} 
q\left(s_{i j}^o | \bm{h}_{i}, \bm{h}_{j}\right) &=\left\{\begin{array}{ll}{\frac{1+s_{ij}^h}{2},} & {s_{i j}^o=1} \\ {\frac{1-s_{ij}^h}{2},} & {s_{i j}^o=0}\end{array}\right.\\ &=\left(\frac{1+s_{ij}^h}{2}\right)^{s_{i j}^o}\left(\frac{1-s_{ij}^h}{2}\right)^{1-s_{i j^o}}, \end{aligned}
\end{equation}
Besides, the weight characterizing class imbalance is measured by $\alpha_{ij}$:

\begin{equation}
\alpha_{i j}=\left\{\begin{array}{l}\frac{\left|\mathcal{S}_{i}\right|\left|\mathcal{S}_{j}\right|}{\sqrt{\left|\mathcal{S}_{i}^{1}\right|\left|\mathcal{S}_{j}^{1}\right|}}, s_{i j}=1 \\ \frac{\left|\mathcal{S}_{i}\right|\left|\mathcal{S}_{j}\right|}{\sqrt{\left|\mathcal{S}_{i}^{0}\right|\left|\mathcal{S}_{j}^{0}\right|}}, s_{i j}=0\end{array}\right.
\end{equation}
where $\mathcal{S}_{i}=\{(i,j)\in \mathcal{E}:\forall j\}$, and

\begin{equation}
\begin{aligned}
&\mathcal{S}_{i}^{1}=\left\{(i, j) \in \mathcal{E}: \forall j, s_{i j}^{o}=1\right\} \\
&\mathcal{S}_{i}^{0}=\left\{(i, j) \in \mathcal{E}: \forall j, s_{i j}^{o}=0\right\} .
\end{aligned}
\end{equation}
The final priority weight is formulated as 
\begin{equation}
w_{i j}=\alpha_{i j}\left(1-q_{i j}\right)^{\gamma},
\end{equation}
where $\gamma$ is a hyper-parameter. With the priority cross-entropy loss, {DPH down-weighs confident image pairs and prioritizes on difficult image pairs with low confidence.} Similarly, priority quantization loss changes the weight for different images to be $w_i'=(1-q_i)\gamma$ and $q_i$ measures how likely a continuous output can be perfectly quantized into {a} binary code. {In this way}, DPH achieved better performance than HashNet.

\emph{Deep Supervised Discrete Hashing} (DSDH)~\cite{li2017deep}. Besides {leveraging the pairwise similarity information}, DSDH also takes advantage of label information by adding a linear regression loss with regularization to the loss function. By {dropping} the binary restrictions, the loss is formulated as:

\begin{equation}\label{eq:dsdh}
\mathcal{L}_{DSDH}=-\sum_{(i,j)\in\mathcal{E}}\left(s_{i j}^o s_{ij}^h-\log \left(1+e^{s_{ij}^h}\right)\right)  +\lambda_1 \sum_{i=1}^{N}||\bm{h}_{i}-sgn(\bm{h}_{i})||_{2}^{2} 
+\lambda_2||\bm{y}_i-\bm{W}^T\bm{b}_i||
+\lambda_3||\bm{W}||_F 
\end{equation}
where $s_{ij}^h=\frac{1}{2}\bm{h}_i^T\bm{h}_j$ and the label is encoded in one-hot format $\bm{y}_i$. {The} second term in Eq. \ref{eq:dsdh} is the linear regression term and the last term is an $\ell_2$ regularization. $\{\bm{h}_i\}_{i=1}^N$, $\{\bm{b}_i\}_{i=1}^N$ and $\bm{W}$ are updated alternatively by using gradient descent method and discrete cyclic coordinate descend method. DSDH greatly increases the performance of image retrieval {since} it takes advantage of both label information and pairwise similarity information. It should be noted that in the linear regression term, the binary code is updated by discrete cyclic coordinate descend, so the constraint of discreteness is met.

\emph{Deep Cauchy Hashing} (DCH) \cite{cao2018deep}. DCH {is} a Bayesian learning framework {similar to} DHN, but it replaced the sigmoid function {with} the function based on Cauchy distribution in the conditional probability. DCH aims to improve the search accuracy with Hamming distances smaller than radius 2. Probability on the basis of generalized sigmoid function could be extremely large when Hamming distances are greater than 2, which is detrimental to current Hamming ball retrieval. DCH tackles this problem by introducing the Cauchy distribution, since the probability drops rapidly if Hamming distances are great than 2. The Cauchy distribution is formulated as 

\begin{equation}
    \begin{aligned}
\sigma\left( d_{ij}^h\right)=\frac{\gamma}{\gamma+ d_{ij}^h},
    \end{aligned}
\end{equation}
{where} $\gamma$ is a hyper-parameter and $d_{ij}^h$ is measured by the normalized Euclidean distance, i.e., $ d_{ij}^h= d(\bm{h}_i,\bm{h}_j)=\frac{L}{2}(1-cos(\bm{h}_i,\bm{h}_j)$. Besides, the prior is based on a variant of the Cauchy distribution, i.e.,

\begin{equation}
    \begin{aligned}
P\left(\bm{h}_{i}\right)=\frac{\gamma}{\gamma+d\left(\left|\bm{h}_{i}\right|, \bm{1}\right)}
    \end{aligned}
\end{equation}
The final loss function is formulated as the log-likelihood plus the quantization loss based on the prior weight. However, this loss function will get almost the same hash code for images with the same label. {Even worse,} the {relationship} for the dissimilar pairs {is} not considered.

\emph{Maximum-Margin Hamming Hashing} (MMHH)~\cite{kang2019maximum}. In view of the shortcomings of DCH, MMHH utilizes the t-Distribution {and contains different objective functions} for similar and dissimilar pairs. The total loss is the weighted sum of two losses. Besides, a margin $\zeta$ is {utilized} to avoid producing the exact same hash codes. The Cauchy distribution in DCH is replaced by: 

\begin{equation}
\sigma\left(d_{i j}^{h}\right)=\left\{\begin{array}{ll}\frac{1}{1+\max \left(0, d_{i j}^{h}-\zeta\right)}, s_{i j}^{o}=1 \\ \frac{1}{1+\max \left(\zeta, d_{i j}^{h}\right)}, \quad s_{i j}^{o}=0\end{array}\right.
\end{equation}
The loss function is the weighted log-likelihood of conditional probability, i.e.,

\begin{equation}
    \begin{aligned}
\begin{aligned} \mathcal{L}_{MMHH} &=\sum_{(i,j)\in\mathcal{E}} w_{i j}\left(s_{i j}^o\right) \log \left(1+\max \left(0, d_{ij}^h-\zeta\right)\right) \\ &+\sum_{(i,j)\in\mathcal{E}} w_{i j}\left(1-s_{i j}^o\right) \log \left(1+\frac{1}{\max \left(\zeta, d_{ij}^h\right)}\right)\\
&+\lambda_1\sum_{i=1}^N ||\bm{h}_i-sgn(\bm{h}_i)||_2^2
 \end{aligned}
    \end{aligned}
\end{equation}
The last term is a standard quantization loss. MMHH also proposed a semi-batch optimization {strategy to alleviate} the imbalance problem. Specifically, the binary codes of the training data are stored as extra memory. The pairwise loss is calculated by the new codes computed in the current epoch and their similar and dissimilar pairs are added into the memory bank for a new epoch. In general, MMHH solves the shortcomings of DCH, which greatly improves search performance.

\emph{Deep Fisher Hashing} (DFH) ~\cite{li2019push}. DFH points out that the pairwise loss minimization is similar to Fisher's Linear discriminant, {which maximizes the gaps between inter-class examples whilst minimizing the gaps between the intra-class examples.} Its logistic loss function is similar to MMHH and the final loss function is formulated as: 
\begin{equation}
 \mathcal{L}_{DFH} =\sum_{(i,j)\in\mathcal{E}} s_{i j}^o \log \left(1+ e^{d_{ij}^h+\epsilon}\right) +\sum_{(i,j)\in\mathcal{E}} \left(1-s_{i j}^o\right) \log \left(1+e^{-d_{ij}^h+\epsilon}\right)
+\lambda_1\sum_{i=1}^N ||\bm{h}_i-sgn(\bm{h}_i)||_2^2,
\end{equation}
where $\epsilon$ is a margin {parameter. Besides, the quantized center loss is added to the objective function, which not only minimizes intra-class distances but also maximizes inter-class distances between binary hash codes of each image.}

\emph{Deep Asymmetric Pairwise Hashing} (DAPH) \cite{shen2017deep}. Similar to ADSH, DAPH also adopted an asymmetric strategy. The difference is that DAPH uses two networks with different parameters for the database and queries. Besides, the bit independence, bit balance and quantization loss are added to the loss function following SH-BDNN. The loss function is optimized {by updating} the two neural networks alternatively.

\emph{Deep Attention-guided Hashing} (DAgH) \cite{yang2019deep}. DAgH adopts a two-step framework {similar to CNNH}, while it utilizes neural networks to learn hash codes in both two steps. {In the first step, the objective function is the combination of the log-likelihood loss and the difference loss with a margin. In the second step, DAgH utilizes binary point-wise cross-entropy for optimization. Besides, the backbone of DAgH includes a fully convolutional network} with an attention module for obtaining accurate deep features.   
          
\emph{Deep Joint Semantic-Embedding Hashing} (DSEH) \cite{li2018deep}. DSEH is the first work to introduce LabNet in deep supervised hashing. {It also adopts a two-step framework with LabNet and ImgNet, respectively.} LabNet is a neural network designed to capture abundant semantic correlation with image pairs, which can help to guide the hash code learning in {the second step.} $\bm{f}_i$ denotes the label embedding produced from one-hot label $\bm{y}_i$. LabNet replaces the input from images to their label and learns the hash codes from labels with a general hashing loss function. {In the second step, ImgNet utilizes an asymmetric loss between the labeled features in the first step} and the newly obtained features from {ImageNet $\bm{h}_j$, i.e. $s_{ij}^h={\bm{f}_i}^T\bm{h}_j $ along with the binary cross-entropy loss similar to} DAgH \cite{yang2019deep}. DSEH fully makes use of the label information from the perspectives of both pairwise loss and cross-entropy loss, which can help generate discriminative and similarity-preserving hash codes. 

\emph{Asymmetric Deep Semantic Quantization} (ADSQ) ~\cite{yang2019asymmetric}. ADSQ increases the performance by utilizing two hashing networks and reducing the difference between the continuous network outputs and the desired hash codes, and the difference loss is also involved. 

\emph{Deep Anchor Graph Hashing} (DAGH) \cite{chen2019deep}. {In the anchor graph, a minimal number of anchors are used to link the whole dataset, allowing for implicit computation of the similarities between distinct examples.} At first, it samples a number of anchors and builds an anchor graph between training samples and anchors. Then the loss function can be divided into two parts. The first part contains a typical pairwise likelihood loss and a linear regression loss. In the second part, the loss is calculated by the distances between training samples and anchors in the same class, and both deep features and binary codes are used to compute the distances. {Besides a general pairwise likelihood loss and a linear regression loss}, DAGH minimizes the distances between deep features of training samples and binary codes of anchors belonging to the same class. This method fully utilizes the remaining labeled data during mini-batch training and helps to obtain efficient binary hash codes.

\subsection{Ranking-based methods}

In this section, we will review the category of deep supervised hashing algorithms that {use the ranking to preserve the similarity structure. Specifically, these methods attempt to preserve the similarity relationships for over two examples that are calculated in the original and Hamming spaces. We further divide ranking-based methods into two groups:
\begin{itemize}
    \item Triplet methods. Due to the ease with which triplet-based similarities could be obtained, triplet ranking losses are popular in deep supervised hashing. These losses attempt to keep the rankings consistent in the Hamming space and the original space for each sampled triplet. For each triplet $(\bm{x}_i,\bm{x}_j,\bm{x}_k)$ {with} $s_{ij}^o>s_{ik}^o$, they usually attempt to minimize a difference loss with margin~\cite{lai2015simultaneous}, i.e., 
    \begin{equation}
    \begin{aligned}
\mathcal{L}_{Triplet}(\bm{h}_i,\bm{h}_j,\bm{h}_k)=max(0,m+d_{ij}^h-d_{ik}^h)
    \end{aligned}
\end{equation}
where $m$ is a margin parameter. Subsequent works introduce the weights based on ranking for each triplet~\cite{zhao2015deep} or utilize the likelihood loss for preserving triplet ranking~\cite{wang2016deep}. The triplet loss can also be combined with the pairwise loss above~\cite{zhang2015bit}. 
    \item List-wise methods. This sub-class usually considers the rankings in the whole dataset rather than in sampled triplet. An example is to optimize ranking-based metrics, i.e., Average Precision and Normalized Discounted Cumulative Gain~\cite{he2018hashing}. Other works utilize the mutual information~\cite{cakir2019hashing} and matrix optimization~\cite{zhou2019angular} for optimizing the hash network from the view of whole datasets. 
    These methods can release the bias during triplet sampling but usually suffer from poor efficiency. 
\end{itemize}
}


\subsubsection{Triplet methods}

\emph{Deep Neural Network Hashing} (DNNH) \cite{lai2015simultaneous}. DNNH modifies the popular triplet ranking objective \cite{norouzi2012hamming} to preserve the relative relationships of samples. To be more precise, given a triplet $(\bm{x}_i,\bm{x}_j,\bm{x}_k)$ {with} $s_{ij}^o>s_{ik}^o$, the ranking loss with margin is formulated as:
\begin{equation}
    \begin{aligned}
\mathcal{L}_{DNNH}(\bm{h}_i,\bm{h}_j,\bm{h}_k)=max(0,1+d_{ij}^h-d_{ik}^h)
    \end{aligned}
\end{equation}
The loss encourages the binary code $\bm{b}_j$ to be closer to the $\bm{b}_i$ than $\bm{b}_k$. By substituting the Euclidean distance for the Hamming distance, the loss function becomes convex, allowing for straightforward optimization:

\begin{equation}
    \begin{aligned}
\mathcal{L}_{DNNH}(\bm{h}_i,\bm{h}_j,\bm{h}_k)=max(0,1+||\bm{h}_i-\bm{h}_j||_2^2-||\bm{h}_i-\bm{h}_k||_2^2)
    \end{aligned}
\end{equation}
Besides, DNNH introduces a sigmoid activation function along with a piece-wise threshold function, which encourage the continuous outputs to approach discrete codes. {The piece-wise threshold function is defined as:}

\begin{equation}
    \begin{aligned}
g(s)=\left\{\begin{array}{lr}{0,} & {s<0.5-\epsilon} \\ {s,} & {0.5-\epsilon \leq s \leq 0.5+\epsilon} \\ {1,} & {s>0.5+\epsilon}\end{array}\right.
   \end{aligned}
\end{equation}
where $\epsilon$ is a small positive hyper-parameter. It is evident that most elements of the outputs will be exact $0$ or $1$ by using this piece-wise threshold function, thus {resulting in} less quantization loss. 

\emph{Deep Regularized Similarity Comparison Hashing} (DRSCH)~\cite{zhang2015bit}. Besides the triplet loss, DRSCH also took advantage of pairwise information by introducing a difference loss as the regularization term. {The bit weights are also included when calculating the distances in the Hamming space.}

\emph{Deep Triplet Supervised Hashing} (DTSH) \cite{wang2016deep}. DTSH replaces the ranking loss by the negative log triplet label likelihood as:
\begin{equation}
    \begin{aligned}
\mathcal{L}_{DTSH}(\bm{h}_i,\bm{h}_j,\bm{h}_k)=log(1+e^{s_{ij}^h-s_{ik}^h-m})-(s_{ij}^h-s_{ik}^h-m),
   \end{aligned}
\end{equation}
which considers the conditional probability \cite{li2017deep} and $m$ is a margin parameter.

\emph{Deep Semantic Ranking-based Hashing} (DSRH) \cite{zhao2015deep}. DSRH {leverages} a surrogate loss based on triplet loss. {Given query ${q}$ and database $\{\bm{x}_i\}_{i=1}^N$, the rankings $\{r_i\}_{i=1}^N$ in database is defined as the number of labels shared with the query. The ranking loss is defined in a triplet form:}

\begin{equation}
    \begin{aligned}
\mathcal{L}_{DSRH}=\sum_{i=1}^N\sum_{j:r_j<r_i} w(r_i,r_j)\delta max(0,\epsilon+d_{qi}^h-d_{qj}^h),
   \end{aligned}
\end{equation}
where $\delta$ and $\epsilon$ are two hyper-parameters {and} $w(r_i,r_j)$ is the weight for each triplet:
\begin{equation}
\omega\left(r_{i}, r_{j}\right)=\frac{2^{r_{i}}-2^{r_{j}}}{Z}
\end{equation}
The form of weights comes from Normalized Discounted Cumulative Gains \cite{jarvelin2017ir} score and $Z$ is a normalization constant which can be omitted. Besides, the bit balance loss and weight regularization are added to the loss function. DSRH improves deep hashing by the surrogate loss, especially on multi-label image datasets.

\subsubsection{Listwise Methods}

 \emph{Hashing as Tie-Aware Learning to Rank} (HALR) \cite{he2018hashing}. HALR explicitly optimizes popular ranking-based assessment metrics including average precision and normalized discounted cumulative gain, which improves the retrieval performance based on ranking. 
 It is noticed that tied ranks may occur due to integer-valued Hamming distance. Hence, HALR introduces a tie-aware formulation of these metrics and trains the hashing network using their continuous relaxations for effective optimization.

\emph{Hashing with Mutual Information} (HashMI) \cite{cakir2019hashing}. HashMI {follows the idea of minimizing neighborhood ambiguity and derives a loss term} based on mutual information, which is sufficiently connected to the aforementioned ranking-based assessment metrics. Given an image $\bm{x}_i$, the random variable $\mathcal{V}_{i,\Phi}$ is defined as a mapping from $\bm{x}_j$ to $d_{ij}^h$, where $\Phi$ is the hashing network. $\mathcal{C}_i$ is the set of images that share the same label with $\bm{x}_i$, i.e., the neighbor of $\bm{x}_i$. The mutual information is defined as 

\begin{equation}
    \begin{aligned}
\mathcal{I}_{HashMI}(\mathcal{V}_{i,\Phi};\mathcal{C}_{i})= H(\mathcal{C}_{i})-H(\mathcal{C}_{i}|\mathcal{V}_{i,\Phi})
    \end{aligned}
\end{equation}
{
 The mutual information is incorporated over the deep feature space for any hashing network $\Phi$, such that a measurement of the quality is obtained which desires to be maximized:
\begin{equation}
    \begin{aligned}
\mathcal{O}=-\int_\Omega \mathcal{I}(\mathcal{V}_{i,\Phi};\mathcal{C}_{i})p_id\bm{x}_i,
    \end{aligned}
\end{equation}
where $\Omega$ is the sample space and $p_i$ denotes the prior distribution which can be removed. After discretion, the loss function turns into: 
\begin{equation}
    \begin{aligned}
\mathcal{L}_{HashMI}= -\sum_{i=1}^N \mathcal{I}(\mathcal{V}_{i,\Phi};\mathcal{C}_{i}),
    \end{aligned}
\end{equation}
whose gradient can be calculated by relaxing the binary constraint and effective minibatch back propagation. The minibatch back propagation is able to effectively retrieve one example against the other example within a minibatch cyclically similar to leave-one-out validation. 
}

\emph{Angular Deep Supervised Hashing} (AnDSH) \cite{zhou2019angular}. AnDSH calculates the Hamming distance between images of different classes to form an upper triangular matrix with size {$K$ by $K$, where $K$ is the number of categorizations.} The mean of Hamming distance matrices is maximized while the variance of the matrices is minimized to make sure that all elements in the matrix could be covered by these hash codes and from the view of bucket theory there is no weakness, i.e., {achieving bit balance.} Besides, this method utilizes classification loss {similar to PCDH} but replaces the softmax loss by A-softmax objective \cite{liu2017sphereface} that could obtain potentially larger inter-class variation along with larger inter-class separation.

\subsection{Pointwise Methods}\label{sec:point-wise}
In this section, {we review pointwise methods that directly take advantage of label information instead of similarity information. Early methods usually add a classification layer to map the hash-like representations into label distributions~\cite{lin2015deep,yang2015supervised,zhang2016efficient,jain2017subic,ma2018multi,zhang2018deep,zhang2019deep}. Then the hash codes are enhanced with the standard classification loss in label space. Further works include the probabilistic models for better binary optimization~\cite{shen2019embarrassingly}.
Recent methods usually build the classification loss in the Hamming space instead. Specifically, they will generate some central hash codes\footnote{They can be also called target codes.}, each of which is associated with a class label. These methods enforce the network outputs to approach their corresponding hash centers with different loss terms, i.e., binary cross-entropy~\cite{yuan2019central}, difference loss~\cite{chen2019hadamard}, polarization loss~\cite{fan} and softmax loss~\cite{hoe2021one} and partial softmax loss~\cite{tu2021partial}. These hash centers are mostly produced by Hadamard matrix~\cite{yuan2019central,chen2019hadamard,hoe2021one}, random sampling~\cite{yuan2019central,hoe2021one} as well as adaptive optimization~\cite{tu2021partial}, which achieves better performance compare with the former two manners.}

\emph{Deep Binary Hashing} (DBH)~\cite{lin2015deep}. {After pre-training of a convolution neural network on the ImageNet, DBH adds a latent layer with sigmoid activation}, where the neurons are utilized to learn hash-like representations while fine-tuning with classification loss on the target dataset. The outputs of the latent layer are discretized into binary hash codes. DBH also emphasizes that the obtained hash codes are for coarse-level search because the {quality} of hash codes is limited. 

\emph{Supervised Semantics-preserving Deep Hashing} (SSDpH)~\cite{yang2015supervised}. SSDpH utilizes a similar architecture to DBH and adds {the quantization loss and the bit balance loss for regularization.} {In this way, SSDpH can produce high-quality hash codes for better retrieval performance.}

\emph{Very Deep Supervised Hashing} (VDSH) \cite{zhang2016efficient}. VDSH builds a very deep hashing network and trains the network with an efficient algorithm layer-wise inspired by alternating direction method of multipliers (ADMM) \cite{boyd2011distributed}. {In virtue of the strong representation ability of the deeper neural network, VDSH can produce better hash codes for effective image retrieval.}

\emph{SUBIC} \cite{jain2017subic}. SUBIC generates structured binary hash codes consisting of the concatenation of several one-hot encoded vectors (i.e. blocks) and obtains each one-hot encoded vector with several softmax functions (i.e., block softmax). Besides classification loss and bit balance regularization, SUBIC utilizes the mean entropy for quantization loss for each block. {SUBIC can also be applied to a range of downstream search tasks including instance retrieval and image classification.}

\emph{Just Maximizing Likelihood Hashing} (PMLR)~\cite{shen2019embarrassingly}. PMLR integrates two dense layers above the top of the hashing network. It utilizes the probability models to parameterize the hashing network for binary constraints. Then PMLR utilizes a classification loss along with a regularization term for better hash code distributions in Hamming space. 


\emph{Central Similarity Quantization} (CSQ) \cite{yuan2019central}. CSQ also utilizes a classification model but in a different way. First, CSQ generates some central hash codes by the properties of a Hadamard matrix or random sampling from Bernoulli distributions, such that the distance between each pair of centroids is large enough. Each label is corresponding to a centroid in the Hamming space and thus each image has its corresponding semantic hash center according to its label. Afterward, the model is trained by the central similarity loss (i.e., binary cross-entropy) with the supervised label information as well as the quantization loss.
{In formulation, 
\begin{equation}
\mathcal{L}_{CSQ}=\sum_{i=1}^{N} \sum_{l=1}^L [\bm{c}_{i,l}\log \bm{h}_{i,l}+(1-\bm{c}_{i,l})\log (1-\bm{h}_{i,l})] + \lambda_1 \sum_{i=1}^N (|| |\bm{h}_i-\bm{1}|-1||_1)
\end{equation}
where $\bm{c}_i \in \{0,1\}^L$ is the hash center generated from labels and $\bm{h}_i\in (0,1)^L$ is the output of the hashing network.}
It is evident that CSQ directly enforces the generated hash codes to approach the corresponding centroids with some relaxations. The core of CSQ is to map the semantic labels into Hamming space to guide hash code learning directly. Thus, samples with comparable labels are converted to similar hash codes, maintaining the global similarities between image pairs and then resulting in effective hash codes for image retrieval.

\emph{Hadamard Codebook-based Deep Hashing}~(HCDH) \cite{chen2019hadamard}. HCDH also utilizes the Hadamard matrix by minimizing the $\ell_2$ difference between hash-like outputs and the target hash codes with their corresponding labels (i.e., Hadamard loss). Different from CSQ, HCDH trains the classification loss and Hadamard loss simultaneously. Hadamard loss can be interpreted as learning the hash centers guided by their supervised labels in $L_2$ norm. Note that HCDH is able to yield discriminative and balanced binary codes for the property of the Hadamard codebook.

\emph{Deep Polarized Network} (DPN)~\cite{fan}. DPN combines metric learning framework with learning to hash and develops a novel polarization loss which minimizes the distance between hash centers and hashing network outputs. {In formulation, 
\begin{equation}
\mathcal{L}_{DPN}=\sum_{i=1}^{N} \sum_{l=1}^L \max \left(\epsilon-\bm{h}_{il} \cdot \bm{c}_{il}, 0\right)
\end{equation}
where $\bm{c}_i \in \{0,1\}^L$ is the hash center and $\bm{h}_i\in (-1,1)^L$ is the output of the hashing network.}
Different from CSQ, the hash centers can be updated after a few epochs. It has been proved that minimizing polarization loss can simultaneously minimize inter-class and maximize intra-class Hamming distances theoretically. {In this way, the hash codes can be easily derived for effective image retrieval.}

{
\emph{OrthHash}~\cite{hoe2021one}. OrthHash is an one-loss model that gets rid of the hassles of tuning the balance coefficients of various losses. Similar to CSQ, OrthHash generates hash centers using Bernoulli distributions. Then, it maximizes the cosine similarity between the hashing network outputs and their corresponding hash centers. In formulation, 
\begin{equation}\label{eq:orth}
\mathcal{L}_{OrthHash}=- \sum_{i=1}^{N} \log \frac{\exp \left(\bm{c}_i^{\top} \bm{h}_{i}\right)}{\sum_{\bm{c}\in \mathcal{C} } \exp \left(\bm{c}^{\top} \bm{h}_{i}\right)}
\end{equation}
where $\mathcal{C}$ denotes the set of all hash centers. Compared with CSQ and DPN, OrthHash not only compare the network outputs and corresponding hash centers, but also considers the other hash centers of different labels. In this way, OrthHash improves the discriminativeness of hash codes. Moreover, since Hamming distance is equivalent to cosine distance for hash codes, OrthHash can promise quantization error minimization. With a single classification objective, it realizes the end-to-end training of deep hashing with promising performance. }  

{\emph{Partial-Softmax Loss based Deep Hashing} (PSLDH)~\cite{tu2021partial}. PSLDH generates a semantic-preserving hash center for each label instead of using Hadamard matrix or random sampling~\cite{yuan2019central}. Specifically, it not also minimizes the inner product of each hash center pair, but also maximizes the information of each hash bit with a bit balance loss term. Moreover, PSLDH trains the hashing network with a partial-softmax loss, which compares the network outputs with both their corresponding hash centers and other centers of partial categories in the datasets. Let $\bm{c}^j$ denote the hash center associated with the $j$-th category. The loss is formulated as:
\begin{equation}
\mathcal{L}_{PSLDH}=\sum_{i=1}^{N} \sum_{j \in \Gamma_{i}}-\log \frac{\exp \left(\eta\left(\bm{h}_{i}^{T} \bm{c}^{j}-\mu L\right)\right)}{\exp \left(\eta\left(\bm{h}_{i}^{T} \bm{c}^{j}-\mu L\right)\right)+\sum_{q \in \Psi_{i}} \exp \left(\eta \bm{h}_{i}^{T} \bm{c}^{q}\right)}
\end{equation}
where $\Gamma_{i}$ denotes the index set of categories associated with  $x_i$, and $\Psi_{i}$ denotes the index set of categories unassociated with $x_i$.} 

\subsection{Quantization}

The quantization techniques have been presented to be derivable from our aforementioned difference loss minimization in Sec.~\ref{sec:pairwise} ~\cite{wang2017survey}. From a statistical standpoint, the quantization error could bound the distance reconstruction error~\cite{jegou2010product}.
As a result, quantization can be used for deep supervised hashing. These methods usually leverage deep neural networks to generate deep features and then adopt product quantization approaches for subsequent quantization. Hence, they optimize the deep features with pairwise difference loss~\cite{cao2016deep}, pairwise likelihood loss~\cite{eghbali2019deep} and triplet loss~\cite{liu2018deep} for better retrieval performance. Further works combine label semantic information for discriminative deep features~\cite{cao2017deep}. Recent works~\cite{klein2019end,eghbali2019deep} integrate deep neural networks into the process of product quantization rather than feature generation and achieve better performance. Other than product quantization, composite quantization can also be enhanced by deep learning~\cite{chen2019similarity}.
Then, we will review the typical deep supervised hashing methods based on quantization.

\emph{Deep Quantization Network} (DQN) \cite{cao2016deep}. DQN generates hash code $b_i$ from the obtained representation $z_i\in \mathbb{R}^D$ with semantics preserved using the product quantization method. First, it decomposes the feature space into the target space, i.e., a Cartesian product of $M$ low-dimensional subspaces and each subspace is quantized into $T$ codewords via clustering. More precisely, the original feature is partitioned into $M$ sub-vectors i.e., $\bm{z}_i=[\bm{z}_{i1};\dots;\bm{z}_{iM}], i=1,\dots,N$ and $\bm{z}_{im}\in \mathbb{R}^{D/M}$ is the sub-vector of $\bm{z}_i$ in the $m$-th subspace. Thus, all sub-vectors in each subspace are quantized into $T$ codewords using K-means without mutual influences. The total loss is defined as follows:
\begin{equation}
\begin{gathered}
\mathcal{L}_{D Q N}=\sum_{i, j} (s_{i j}^{o}-\cos \left(z_{i}, z_{j}\right))^2+\lambda_{1} \sum_{m=1}^{M} \sum_{i=1}^{N}\left\|z_{i m}-C_{m} b_{i m}\right\|_{2}^{2} \\
s.t. \left\|b_{i m}\right\|_{0}=1, b_{i m} \in\{0,1\}^{T}
\end{gathered}
\end{equation}
where $cos(\cdot)$ denotes the cosine similarity metric and $\bm{C}_m=[\bm{c}_{m1},\dots,\bm{c}_{mT}]$ represents $T$ codewords of the $m$-th subspace, and $\bm{b}_{im}$ is the one-hot embedding to guide which codeword in $\bm{C}_m$ should be used to approach the $i$-th point $\bm{z}_{im}$. 
Mathematically, the second term, i.e., product quantization can be reformulated as:
\begin{equation}
    \begin{aligned}
    \sum_{i=1}^N||\bm{z}_i-\bm{C}\bm{b}_i||_2^2,
    \end{aligned}
\end{equation}
where $\bm{C}$ is a $D\times MT$ matrix can be written as 
$\bm{C}=diag(\bm{C}_1,\dots,\bm{C}_M)$.
Note that the quantization loss of converting the feature $\bm{z}_i$ into binary code $\bm{b}_i$ can be restricted via minimizing $Q$. Besides, quantization-based hashing also adds pairwise similarity preserving loss to the final loss function. Finally, Asymmetric Quantizer Distance (AQD) is widely used for approximate nearest neighbor search, which is formulated as:
\begin{equation}
    \begin{aligned}
AQD(\bm{q},\bm{x}_i)=\sum_{m=1}^M||\bm{z}_{qm}-\bm{C}_m\bm{b}_{im}||_2^2,
    \end{aligned}
\end{equation}
where $\bm{z}_{qm}$ is the $m$-th sub-vector for the feature of query $\bm{q}$.

\emph{Deep Triplet Quantization} (DTQ)~\cite{liu2018deep}. DTQ uses a triplet loss to preserve the similarity information and a smooth orthogonality regularization is added to the codebooks which are similar to the bit independence. {Let $\mathcal{T}$ denote the set of all triplet. Each triplet $(\bm{x}_i,\bm{x}_j,\bm{x}_k)$ satisfies  $s_{ij}^o>s_{ik}^o$. The total loss function is as follows:
\begin{equation}
\begin{aligned}
 \mathcal{L}_{DTQ} =& \sum_{(\bm{x}_i,\bm{x}_j,\bm{x}_k)\in \mathcal{T}} (\max(0, \epsilon+||\bm{z}_i- \bm{z}_j ||_2^2-||\bm{z}_i- \bm{z}_k ||_2^2)+\lambda_{1} \sum_{m=1}^{M} \sum_{i=1}^{N}\left\|z_{i m}-C_{m} b_{i m}\right\|_{2}^{2}  \\
&+\lambda_2 \sum_{m=1}^M\sum_{m'=1}^M||\bm{C}_m^T\bm{C}_{m'}-\bm{I}||^2.
\end{aligned}
\end{equation}
The last term is the orthogonality penalty term.  }
In addition, DTQ selects triplets by Group Hard to make sure that the number of explored valid triplets is suitable for optimization. Specifically, the training data is split into various groups, and a hard (i.e., with positive triplet loss) negative example is picked randomly as an anchor-positive image pair from every group.

\emph{Deep Visual-semantic Quantization} (DVsQ)~\cite{cao2017deep}. DVsQ optimizes the quantization network using labeled image samples along with the semantic messages from their latent text domains. Specifically, by using the image representations $\bm{z}_i$ from the pre-trained network, it produces deep visual-semantic representations.
They are then trained to forecast the word embeddings $\bm{v}$ (i.e., $\bm{v}_i$ for label $i$), which are further estimated by a skip-gram model. The loss function includes the adaptive margin ranking loss and a quantization loss:
\begin{equation}
    \begin{aligned}
\mathcal{L}_{DVsQ}=\sum_{i=1}^N\sum_{j\in \bm{y}_i}\sum_{k\notin \bm{y}_i}max(0,\delta_{jk}-cos(\bm{v}_j,\bm{z}_i)+cos(\bm{v}_k,\bm{z}_i)) + \lambda_1 \sum_{i=1}^N\sum_{j=1}^{|\bm{y}|}||\bm{v}_j^T (\bm{z}_i-\bm{C}\bm{b}_i)||_2^2.
    \end{aligned}
\end{equation}
where $\bm{y}_i$ is the label set of the $i$-th image, and $\delta_{jk}$ is an adaptive margin and the quantization loss is inspired by the maximum inner-product search. 
DVsQ adopts the same strategy as LabNet and combines the visual information and semantic quantization in a uniform framework instead of a two-step approach. By this means, DVsQ greatly improves the retrieval performance. 

\emph{Deep Product Quantization} (DPQ) \cite{klein2019end}. DPQ leverages both the powerful capacity of product quantization (PQ) and the end-to-end learning ability of deep learning to optimize the clustering results of product quantization through classification tasks. 
Specifically, for each input $\bm{x}_i $, it first uses an embedding layer and an MLP to obtain the deep representation $\bm{z}_i \in \mathbb{R}^{MF}$. Then the representation is sliced into $M$ sub-vectors with $z_{i,m} \in \mathbb{R}^F$ similar to PQ. Different from DQN, an MLP is used to turn each sub-vector into a probabilistic vector with $T$ elements $p_m(t), t=1,\dots,T$ by softmax function. The matrix $\bm{C}_m\in \mathbb{R}^{T\times D}$ denotes the $T$ centroids. $p_m(k)$ denotes the probability that the $m$-th sub-vector is quantized by the $t$-th row of $\bm{C}_m$. The soft representation of the $m$-th sub-vector is calculated by combining the row vectors of $\bm{C}_m$. 
\begin{equation}
soft_m=\sum_{t=1}^Tp_m(t)\bm{C}_m(t).
\end{equation}
Considering the probability $p_m(k)$ in one-hot format, given $t^*=argmax_t p_m(t)$, the hard probability is denoted as $e_m(t)=\delta_{tt*}$ in one-hot format and we have:
\begin{equation}
hard_m=\sum_{t=1}^Te_m(t)\bm{C}_m(t).
\end{equation}
The obtained sub-vectors of soft and hard representations are then concatenated to produce the ultimate representations, i.e., $\text{soft}=[\text{soft}_1, \dots, \text{soft}_M]$ and $\text{hard}=[\text{hard}_1, \dots, \text{hard}_M] \in \mathbb{R}^{MD}$. Each representation is followed by a fully-connected classification layer. Besides two classification losses, the joint central loss is also added by first learning the center vector for each categorization and minimizing the distances between deep features. It is noticed that both the soft and hard representations come from the same centers in DPQ, which encourages both representations to approach the centers, reducing the disparity between the soft and hard representations.
This helps to improve the discriminative power of the features and to contribute to the retrieval performance. Gini batch loss and Gini sample loss are also introduced for the class balance and encourage the two representations of the same image to be closer. {Overall, DPQ replaces the k-means process in PQ and DQN technique with deep learning combined with a classification model and is able to create compressed representations for fast classification and fast image retrieval. }

\emph{Deep Spherical Quantization} (DSQ) \cite{eghbali2019deep}. {DSQ first uses the deep neural network to obtain the $\ell_{2}$ normalized features and then quantizes these features on a unit hypersphere with an elaborate quantization manner.}
After constraining the continuous representations to staying on a unit hypersphere, DSQ attempts to reduce the reconstruction loss using multi-codebook quantization (MCQ). Different from PQ, MCQ draws near the representation vectors with the summation of multiple codewords instead of the concatenation. $\hat{\bm{y}}_i$ denotes the predicted label distribution. $\phi_{y_i}$ denotes the feature center of the $y_i$-th class.  
The overall loss for training the model is as follows:
\begin{equation}
    \begin{aligned}
\mathcal{L}_{DSQ}=& \sum_{i=1}^N -\log \bm{y}_i^T log \hat{\bm{y}}_i+\lambda_1 \sum_{i=1}^N||\bm{z}_{i}-[\bm{C}_1,\dots,\bm{C}_M]\bm{b}_{i}||_2^2\\
&+ \lambda_2 \sum_{i=1}^N ||\bm{z}_i - \phi_{y_i} ||_2^2 + \lambda_3 \sum_{i=1}^N ||\phi_{y_i}-\bm{C}\bm{b}_i||_2^2\\
& s.t. ||\bm{b}_{im}||_0=1, \bm{b}_i \in\{0,1\}^K, \bm{b}_i=[\bm{b}_{i,1}^T,\dots,\bm{b}_{i,M}^T]^T. 
    \end{aligned}
\end{equation}
where the first, second, third and last term is the softmax loss, quantization loss, the center loss and the discriminative loss, respectively.
The last two losses encourage both the quantized vectors and deep features to approach their centers, respectively.

\emph{Similarity Preserving Deep Asymmetric Quantization} (DPDAQ) \cite{chen2019similarity}. DPDAQ adopts Asymmetric Quantizer Distance to approach the desired similarity metric, which is similar to ADSH. {Moreover, it uses composite quantization instead of product quantization and the representations in the training set come from the deep neural network in an unquantized form.} SPDAQ also takes advantage of similarity information and label information to achieve better retrieval performance.

\subsection{Other Techniques for Deep Hashing}
\subsubsection{Hashing with Generative Adversarial Networks}
Generative Adversarial Networks (GANs) \cite{goodfellow2014generative} are popular neural network models to generate virtual examples without needing supervised knowledge. {There are also several hashing methods leveraging GANs to enhance the performance.}

\emph{Deep Semantic Hashing with GAN} (DSH-GAN) \cite{qiu2017deep}. DSH-GAN is the first hashing method that takes advantage of GANs for image retrieval. It typically includes four components, i.e., a neural network to produce image representations, an adversarial discriminator for differentiating between synthetic images and real images, a hashing network for projecting representations into binary codes and a classification head. Specifically, the generator network attempts to generate synthetic images using the concatenation of the label embedding and generated noise vector. The discriminator attempts to jointly differentiate between real samples and synthetic ones and categorize the inputs into proper semantic labels. Finally, the overall framework is optimized using the adversarial loss to mix two sources and the classification loss to obtain the ground truth labels using a classic minimax mechanism. The input of the network is image triplets, each of which contains three images. The first one is a real image treated as a query, the second one is a synthetic image created with the same label as the query image by the generator network, and the third one is a synthetic image with different semantics. GAN provides a hashing model with strong generalization potential from the maintaining of semantics and similarity, which improves the quality of hash codes.

\emph{HashGAN}~\cite{cao2018hashgan}. {HashGAN augments the training data with images synthesized by pair conditional Wasserstein GAN (WGAN) inspired by \cite{gulrajani2017improved} which sufficiently explores the pairwise semantic relationships.} In this module, the training samples along with the pairwise similarities are considered as inputs and a generator and a discriminator is trained simultaneously by adding the pairwise similarity besides the loss function of WGAN. The hash encoder 
produces high-quality binary codes for all occurred pictures using a likelihood objective similar to HashNet. HashGAN is also capable of coping with the dataset without class labels but with {pairwise} similarity information.

\subsubsection{Ensemble Learning}
 Guo et al.~\cite{guo2018trivial} point out that for the current deep supervised hashing model, simply increasing the length of the hash code with a single hashing model cannot significantly enhance the performance. The potential cause is that the loss functions adopted by existing methods are prone to produce highly correlated and redundant hash codes. 
 Inspired by this, several methods attempt to leverage ensemble learning to increase the retrieval performance with more hash bits.

\emph{Ensemble-based Deep Supervised Hashing} (EbDSH)~\cite{guo2018trivial}. {EbDSH leverages an ensemble learning strategy for better retrieval performance.} Specifically, it trains a number of deep hashing models with different training datasets, training data, initialization and networks, then concatenates them into the final hash codes. It is noticed that the ensemble strategy is suitable for parallelization and incremental learning.  

\emph{Weighted Multi-deep Ranking Supervised hashing} (WMRSH) \cite{li2019weighted}. {WMRSH attempts to generate a high-quality hash function using multiple hash tables derived from the hashing networks.} To be specific, WMRSH adds bit-wise weights and table-wise weights for each bit in each hash table. For each bit in a table, the similarity preservation is measured by product loss. Afterward, the bit independence is measured by the correlation between two bits. Finally, the table-wise weight can be derived from the mean average precision for every hash table. The final weight is the product of the three above terms for the final hash codes (i.e., the concatenation of the hash tables with weights). A similar strategy called Hash Boosting has been introduced in \cite{liu2019mutual}. 

Apart from these methods, NMLayer~\cite{fu2019neurons} balances the importance of each bit and merges the redundant bits together to learn more compact hash codes.

\subsubsection{Training Strategy for Deep Hashing}
In this subsection, we will introduce two methods that adopt different training strategies from most other methods.

\emph{Greedy Hash} \cite{su2018greedy}. Greedy Hash adopts a greedy algorithm for fast processing of hashing discrete optimization by introducing a hash layer with a sign function instead of the quantization error. {To overcome the ill-posed gradient problem~\cite{yang2018semantic}, the gradients are transmitted entirely to the front layer, which effectively prevents the vanishing gradients of the sign function and updates all bits together. This strategy is also adopted in recent works~\cite{qiu2021unsupervised}.}
  
{\emph{Gradient Attention deep Hashing} (GAH) \cite{huang2019accelerate}. This work points out a dilemma in learning deep hashing models through gradient descent that it makes no difference to the loss if the paired hash codes change their signs together.}
As a result, GAH generates attention on the derivatives of each hash bit for each image by maximizing the decrease of loss during optimization. It leverages a gradient attention network with two fully-connected layers to produce normalized weights and then applies them to the derivatives in the last layer. In conclusion, this model optimizes the training process by adopting a gradient attention network for acceleration.

\begin{table}

\renewcommand{\arraystretch}{1.3}
\tiny
\caption{{A Summary of Deep Unsupervised Hashing Approaches w.r.t the Manner of Generating Similarity Information, Generating and Handling the Pseudo-Label, Binarization as well as Other Skills. Drop = Drop the sign operator in the neural network and treat the binary code as an approximation of the network output, Reg. = Regression, Quan. = Quantization, Dist. = Distance, Conf. = Confidence, Trans. = Transformation, Ind. = Independence, Bal. = Balance., Cla. = Classification, Clu. = Clustering. }}
\label{table_unsupervised}
\centering
\begin{tabular}{l|c|c|c|c}
\hline
 Approach  & Similarity Information & Pseudo-Label  & Binarization & Other skills\\
\hline
            SSDH \cite{yang2018semantic}   & Local Dist. & - & Tanh & -   \\
            DistillHash \cite{yang2018semantic}  & Local Dist.+Neighbour Information & - & Tanh & -   \\
            SADH~\cite{shen2018unsupervised}  & Network Output + Adjacent Matrix & - & Drop + Alternation & -   \\
            MLS$^3$RUDH~\cite{tu2020mls3rduh}   & Local Dist. + Manifold Dist. & - & Tanh & -   \\
            TBH~\cite{shen2020auto}   & Hash Codes  & - &  Bottleneck Reg. &  AE   \\
            GLC \cite{luo2021deep}   & Local Dist.+ K-Means & -  & Tanh & -   \\
            MBE~\cite{li2021deep} & Local Dist. & - & -  & Bit Bal. with Bi-Half Layer \\
            CIMON \cite{luo2020cimon}   & Local Dist.+ Spectral Clu. + Conf. & - & Tanh & Contrastive Learning   \\
            DATE \cite{luo2021mm}   & Local Dist. + Distribution Dist. +Conf. & - & Tanh & Contrastive Learning   \\
            PLUDDH~\cite{hu2017pseudo}   & - & Kmeans + Cla. Layer & Tanh + Quan. Loss & -   \\
            DAVR~\cite{huang2016unsupervised}   & - & Deep Clu. + Triplet Loss & Drop & -   \\
            CUDH~\cite{gu2019clustering}  & - & Deep Embedding Clu. & Tanh+ Quan. Loss & -   \\
            DVB~\cite{shen2019unsupervised} & Adjacent Matrix & Clu. & Quan. Loss & VAE + Bit Indep. \\
        
            UDHPL \cite{zhang2017unsupervised} & - & Kmeans + PCA + MI & - & - \\
            DU3H~\cite{zhang2020deep} & Local Dist. + Conf. & Kmeans + Hash Center & Tanh & GCN \\
            UDMSH~\cite{qin2020unsupervised} & Local Dist. + Conf. & - & Quan. Loss & - \\
             DSAH~\cite{lin2021deep} & Updated Local Dist.   + Conf. & - & Quan. Loss & - \\
            UDKH~\cite{dong2020unsupervised}   & - & Hash Code + Deep Clu. & Alternative & -   \\
            BDNN~\cite{do2016learning} & -  & - & Drop & AE \\
            DH~\cite{erin2015deep}   & - & - & Quan. Loss  & Bit Bal. + Ind.  \\
            DeepBit~\cite{lin2016learning}   & - & - & Quan. Loss  & Bit Bal. + Trans. Reg.  \\
            UTH~\cite{lin2016learning}   & - & - & Quan. Loss  & Bit Bal. + Triplet Trans. Reg.  \\
            BGAN~\cite{song2018binary} & Local Dist. & - & Tanh + Continuation & GAN \\ 
            BinGAN~\cite{zieba2018bingan}  & - & - & Quan. Loss & GAN + Bit independence \\
            HashGAN~\cite{ghasedi2018unsupervised}   & - & - & Quan. Loss  & Bit Bal. + Ind. +  Trans. Reg. + GAN   \\
            SGH~\cite{dai2017stochastic}   & - & - & -  & VAE  \\
            CIBHash~\cite{qiu2021unsupervised}   & - & - & Drop  & Contrastive Learning \\
            SPQ~\cite{jang2021self}   & - & - & Quantization  & Cross Contrastive Learning  \\
            HashSIM~\cite{luo2022improve}   & Local Dist. & - & Tanh  & Bit Contrastive Learning  \\

\hline

\end{tabular}
\end{table}

\section{Deep Unsupervised Hashing}

\subsection{Overview}

{Recently, unsupervised hashing methods have received widespread attention due to their sufficient leverage of the unlabeled data, which facilities the practical applications in the real world. Since deep unsupervised methods can not acquire label information, the semantic information is obtained in deep feature space with pre-trained networks. With semantic information, the problem can be converted into a supervised problem. However, how to infer semantics information and how to utilize semantics information for learning hash codes are two key problems here.
According to semantics learning manners, the unsupervised methods can be mainly classified into three categories, i.e., similarity reconstruction-based methods, pseudo-label-based methods and prediction-free self-supervised learning-based methods. Similarity reconstruction-based methods usually generate pairwise semantic information, and then leverage pairwise semantic preserving techniques in Sec. \ref{sec:pairwise} for hash code learning. Pseudo-label-based methods usually produce pointwise pseudo-labels for inputs and then leverage pointwise semantic preserving techniques in Sec. \ref{sec:point-wise} for hash code learning. Lastly, prediction-free self-supervised learning-based methods leverage data itself for training without generating explicit semantic information, i.e., similarity signals and pseudo-labels.
Specifically, they usually utilize regularization terms, auto-encoder models, generative models and contrastive learning to produce high-quality hash codes. The regularization terms include bit balance loss term, bit independence loss term and a transformation-invariant regularization term. Several approaches may combine different kinds of semantics learning manners. The optimization of binarization is still an important problem for deep unsupervised hashing. Most of the methods use $tanh(\cdot)$ to approximate $sign(\cdot)$ and generate approximate hash codes by the hashing network for optimization. 
The summary of these algorithms is shown in Table \ref{table_unsupervised}. Then, we elaborate on these classes as below.}

\begin{figure*}[t]
    \centering
    \includegraphics[width=10cm,keepaspectratio=true]{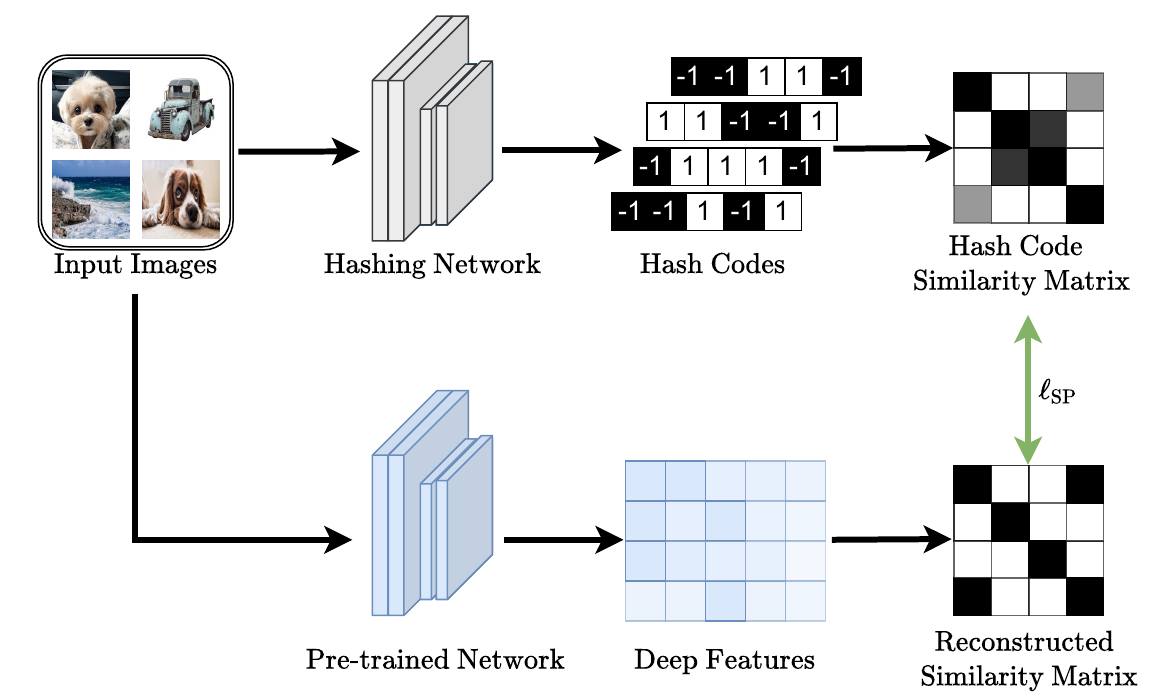}
    \caption{{Basic Framework of Deep Unsupervised Hashing with Similarity Reconstruction. The deep features are extracted by a pre-trained network for similarity reconstruction. More details will be discussed in Sec. \ref{sec:sbm}.}   }
    \label{fig:duh}
\end{figure*}

\subsection{Similarity Reconstruction-based Methods}\label{sec:sbm}
{Similarity reconstruction-based methods aim to leverage pairwise methods to solve the problem. However, the similarity information is unavailable without label annotation. Hence, these methods utilize a two-step framework as shown in Fig. \ref{fig:duh}. Firstly, they extract deep representations $\bm{z}_i$ using the pre-trained neural network and then infer the similarity information $\{s_{ij}^o\}_{(i,j)\in \mathcal{E}}$ by distance metrics in deep feature space. Secondly, a hashing network is trained to create similarity-preserving binary codes by leveraging the reconstructed similarity structure as guidance. With similarity information, the problem can be solved with pairwise supervised methods. The key to this kind of methods is how to generate accurate similarity information. Early methods usually truncate pairwise distances in deep feature space~\cite{yang2018semantic}. Further studies utilizes the neighbourhood information~\cite{yang2019distillhash,luo2021deep,luo2020cimon}, confidence degree~\cite{luo2020cimon}, other similarity matrices~\cite{tu2020mls3rduh,luo2021mm} to obtain a precise similarity structure for reliable guidance of subsequent optimization. Recently, a few researchers argue that static similarity structure from the pre-trained network is not optimal and propose to update it based on obtained hash codes~\cite{shen2020auto,shen2018unsupervised,lin2021deep}. Next, we revise these methods in detail.  
}

{\emph{Semantic Structure-based Unsupervised Deep Hashing} (SSDH) \cite{yang2018semantic}. SSDH is the first study along this line, which applies VGG-F model to extract deep features and perform hash code learning. It studies the cosine distance for each pair in deep feature space, and finds that the distribution of cosine distances can be approximated by two half Gaussian distributions. Hence, through parameter estimation, SSDH sets two distance threshold $d_l$ and $d_r$ and construct a similarity structure as follows:
\begin{equation}\label{eq:ssdh}
s_{ij}^o=\left\{\begin{array}{ll}1, & \text { if } d(\bm{z}_i, \bm{z}_j) \leq d_{l} \\ 0, & \text { if } d_{l}<d(\bm{z}_i, \bm{z}_j)<d_{r} \\ -1, & \text { if } d(\bm{z}_i, \bm{z}_j) \geq d_{r}\end{array}\right.
\end{equation}
where $d(\cdot, \cdot)$ denotes the cosine distance of two vectors. From Eq. \ref{eq:ssdh}, SSDH considers sample pairs with distance smaller than $d_l$ as semantically similar while considers sample pairs with distances large than $d_r$ as semantically dissimilar. Similar to SH-BDNN, a similarity difference loss is adopted as follows:
\begin{equation}
 \mathcal{L}_{SSDH}=\sum_{i=1}^{N} \sum_{j=1}^{N}\left|s_{ij}^o\right|\left(s_{ij}^h -s_{ij}^o\right)^{2}
\end{equation}
where $s_{ij}^h = \bm{h}_i^T\bm{h}_j/L $, $\bm{h}_i$ denotes the output of the deep network with activation function $tanh(\cdot)$. The activation function $sign(\cdot)$ is utilized instead during evaluation. However, the performance of SSDH is limited due to two issues. On the one hand, its similarity structure is typically unreliable using two coarse thresholds. On the other hand, it discards a range of signals in similarity structure.} 

{\emph{DistillHash}~\cite{yang2019distillhash}. DistillHash leverages the similarity signals from local structures to distill similarity signals. Specifically, for each pair of images, it studies the similarities of their neighbors and then removes the similarity signal if it has huge variants in local structures. The distillation process can be implemented with Bayes optimal classifier. Finally, DistillHash leverages likelihood loss minimization to train the hashing network with the similarity structure: 
\begin{equation}
 \mathcal{L}_{DistillHash}=- \sum_{i=1}^{N} \sum_{j=1}^{N} (\bm{1}_{s_{ij}^o=1}  \sigma(s_{ij}^b)+  \bm{1}_{s_{ij}^o=-1} (1-\sigma(s_{ij}^b)))
\end{equation}
The improvement of DistillHash over SSDH is mainly the introduction of local structures to distill confident signals, which releases the first issue in the last paragraph. }

{\emph{Similarity Adaptive Deep Hashing} (SADH)~\cite{shen2018unsupervised}. SADH trains the model alternatively over three parts. In part one, it trains the hashing network under the guidance of binary codes. In part two, it leverages the network output to update the similarity structure. In part three, the hash codes are optimized with network output following the ADMM process. The alternative optimization improves the robustness of the model and helps achieve better hash codes for image retrieval.} 

{\emph{Deep Unsupervised Hashing via Manifold based Local Semantic Similarity Structure Reconstructing}
(MLS$^3$RUDH)~\cite{tu2020mls3rduh}. MLS$^3$RUDH incorporates the manifold structure in deep feature space to generate an accurate similarity structure. Specifically, it leverages a random walk on the nearest neighbor graph to measure the manifold similarity. The final similarity structure is denoted as follows:
\begin{equation}
s_{ij}^o=\left\{\begin{array}{ll}1, & \bm{x}_{j} \in N^{c}\left(\bm{x}_{i}\right) \wedge \bm{x}_{j} \in N^{m}\left(\bm{x}_{i}\right) \\ -1, & \bm{x}_{j} \in N^{c}\left(\bm{x}_{i}\right) \wedge \bm{x}_{j} \notin N^{m}\left(\bm{x}_{i}\right) \\ 0, & \text { otherwise }\end{array}\right.
\end{equation}
where $N^c(\cdot)$ and $N^m(\cdot)$ denote the set of the neighbour samples in terms of both cosine similarity and manifold similarity, respectively. Then the hashing network is optimized through difference loss minimization as:
\begin{equation}
\mathcal{L}_{MLS^3RUDH}= \sum_{i=1}^{N} \sum_{j=1}^{N} \log \left(\cosh \left(s_{ij}^h -s_{ij}^o\right)\right)
\end{equation}
MLS$^3$RUDH leverages the manifold similarity to generate a more accurate similarity structure, which guides the optimization of the hashing network effectively.}

{\emph{Auto-Encoding Twin-Bottleneck Hashing} (TBH)~\cite{shen2020auto}. TBH introduce an adaptive code-driven graph to guide hash code learning. It contains a binary bottleneck to construct code-driven similarity graph and a continuous bottleneck for reconstruction. To be specific, the similarity structure is defined by hash codes:
\begin{equation}
    s_{ij}^o = 1- d_{ij}^h/L
\end{equation}
where $d_{ij}$ is the Hamming distance between $b_i$ and $b_j$. The outputs of the continuous bottleneck are fed into graph neural networks with the similarity structure as the adjacency for the final reconstruction. Moreover, TBH involves adversarial learning to regularize the network for high-quality hash codes. TBH utilizes a dynamic graph guided with the reconstruction loss for accurate similarity structures, which helps hash code preserve better similarity for reliable retrieval.}

{\emph{Deep Unsupervised Hashing by Global and Local Consistency} (GLC) \cite{luo2021deep}. GLC extracts semantic information from both local and global views. For local views, it builds reliable graphs and penalty graphs based on the cosine distances of image pairs. For global views, it utilizes global clustering to derive cluster centers for different classes. During the optimization of hashing network, GLC preserves the local similarity using a product loss and minimizes the Hamming distances between the hash codes in the same cluster. Compared with previous methods, GLC preserves the similarity from different views in a unified manner, resulting in effective retrieval performance. }

{\emph{CIMON} \cite{luo2020cimon}. CIMON first sets a threshold $d_t$ to partition the local similarity signals based on cosine metric into two groups. Inspired by the fact that the representations of samples with the similar semantic information ought to be on
a high-dimensional manifold, CIMON adopts the results of spectral clustering to remove contradictory results for refining the semantic similarities. Moreover, it constructs the confidence of the similarity signals. The semantic information includes similarity signals $\{s_{ij}^o\}_{(i,j)\in \mathcal{E}}$ and their confidence $\{w_{ij}\}_{(i,j)\in \mathcal{E}}$. In formulation, 
\begin{equation}
s_{ij}^o= \begin{cases}1 & c_{i}=c_{j} \& d(\bm{z}_i,\bm{z}_j)<d_{t} \\ -1 & c_{i} \neq c_{j} \& d(\bm{z}_i,\bm{z}_j)<d_{t} \\ 0 & \text { otherwise }\end{cases}
\end{equation}
where $\{c_i\}_{i=1}^N$ is the cluster label of clustering.
The confidence is built based on the cumulative distribution function:
\begin{equation}
w_{i j}= \begin{cases}\frac{\Phi_{1}(d_t)-\Phi_{1}\left( d(\bm{z}_i,\bm{z}_j)\right)}{\Phi_{1}(d_t)-\Phi_{1}(0)} &  d(\bm{z}_i,\bm{z}_j) \leq d_t \& s_{ij}^o \neq 0 \\ \frac{\Phi_{2}\left( d(\bm{z}_i,\bm{z}_j)\right)-\Phi_{2}(d_t)}{\Phi_{2}(2)-\Phi_{2}(d_t)} & d_t< d(\bm{z}_i,\bm{z}_j) \& s_{ij}^o \neq 0 \\ 0 & \hat{S}_{i j}=0\end{cases}
\end{equation}
where $\Phi_\cdot(\cdot)$ is cumulative distribution function of estimated Gaussian distribution. 
CIMON generates two groups of semantic information by data augmentation and matches the hash code similarity with similarity information in a parallel and cross manner. Moreover, contrastive learning is also introduced to improve the quality of hash codes. To our knowledge, CIMON is the first method using contrastive learning for hash code learning and achieves impressive performance due to both reliable similarity information and contrastive learning. }

{\emph{Maximizing Bit Entropy} (MBE)~\cite{li2021deep}. MBE utilizes the continuous cosine similarity signals to guide hash code learning. More importantly, it introduces a bi-half layer for better quantization. Specifically, for the continuous network outputs, MBE sorts the elements of each dimension over all the minibatch samples, and then assigns the top half elements to $1$ and the remaining elements to $-1$. In this manner, MBE can achieve absolute bit balance. The optimization of the bi-half layer is based on a straight-through estimator similar to \cite{su2018greedy}.} 

{\emph{DATE} \cite{luo2021mm}. DATE characterizes each image by a set of its augmented views, which can be considered as examples from its latent distributions. Then it calculates the semantic distances between sample pairs by computing the distribution divergence using a non-parametric way. Specifically, we define the smoothed ball divergence statistic written as: 
\begin{equation}
\begin{aligned}
B D\left(\left\{\bm{z}_{i}^{r}\right\}_{r=1}^{R},\left\{\bm{z}_{j}^{r}\right\}_{m=1}^{R}\right) &=\frac{1}{R} \sum_{r=1}^{R}\left(\left(\frac{1}{R} \sum_{r=1}^{M} d\left(\bm{z}_{i}^m, \bm{z}_{j}^r\right)-d\left(\bm{z}_{i}^m, \bm{z}_{i}^r\right)\right)^{2}\right.\\
&\left.+\left( \frac{1}{R} \sum_{r=1}^{M} d\left(\bm{z}_{j}^m, \bm{z}_{j}^r\right)-d\left(\bm{z}_{j}^m, \bm{z}_{i}^r\right)\right)^{2}\right)
\end{aligned}
\end{equation}
where $\left\{\bm{z}_{i}^{r}\right\}_{r=1}^{R}$ and  $\left\{\bm{z}_{j}^{r}\right\}_{r=1}^{R}$ denotes the features of augmented views of images $\bm{x}_i$ and $\bm{x}_j$ through a pre-trained network. 
Then the distribution distance is combined with cosine distance to generate reliable semantic information. Contrastive learning is also utilized for high-quality hash codes. Through accurate semantic information enhanced by augmentations, DATE can achieve promising performance for image retrieval.}

\subsection{Pseudo-label-based Methods} 

{The second class of deep unsupervised methods generates pseudo-labels. These methods treat pseudo-labels as semantic information and convert this problem into supervised hashing. Most of them first leverage clustering (e.g., K-means and spectral clustering) to generate pseudo-labels~\cite{hu2017pseudo,huang2016unsupervised,zhang2017unsupervised,zhang2020deep,shen2019unsupervised}. Then, these pseudo-labels guide hash code learning with deep supervised hashing methods. Further studies utilize a deep clustering framework to combine clustering with the hashing network to adaptively update pseudo-labels~\cite{gu2019clustering,dong2020unsupervised}. 
}

{\emph{Pseudo Label-based Unsupervised Deep Discriminative Hashing} (PLUDDH)~\cite{hu2017pseudo}. PLUDDH utilizes the pre-trained network to extract deep features and then generates pseudo-labels via clustering. Then the hashing network is supervised by pseudo-labels. It has the same neural network architecture as DBN and trains it with the classification loss and the quantization loss. PLUDDH explores deep feature space with coarse clustering, which may generate false pseudo-labels. Hence, its retrieval performance is limited when the dataset is complicated.}

{\emph{Unsupervised Learning of Discriminative Attributes and Visual Representations} (DAVR)~\cite{huang2016unsupervised} DAVR adopts a two-step framework. In the first stage, a CNN is trained coupled with unsupervised discriminative clustering \cite{singh2012unsupervised} to generate the cluster membership. In the second stage, cluster membership is utilized as supervision to uncover common cluster properties while optimizing their separability using a triplet objective. In general, the unsupervised hashing is converted into a supervised problem by the obtained pseudo labels.}

{\emph{Unsupervised Deep Hashing with Pseudo Labels} (UDHPL) \cite{zhang2017unsupervised}. UDHPL first extracts features and reduces their dimension with Principle Component Analysis (PCA) to release the noise. Then it generates the pseudo-labels through the Bayes' rule. UDHPL maximizes the correlation between the projection vectors of pseudo-labels and deep features, and the features can be projected into the Hamming space. With a rotation matrix, the hash code can be generated, which will guide the optimization of the hashing network. UDHPL improves the pseudo-labels through PCA and guides the network training with mutual information maximization, which helps to preserve similarity information for effective retrieval. }

{\emph{Clustering-driven Unsupervised Deep Hashing} (CUDH)~\cite{gu2019clustering}. CUDH first extracts deep features from the pre-trained network. Inspired by the deep clustering model DEC~\cite{xie2016unsupervised} which performs clustering in the embedding space, it modifies the model to iteratively learn discriminative clusters in the Hamming space with extra quantization loss. 
CUDH is capable of generating discriminative hash code in virtue of the deep clustering model. }

{\emph{Deep Unsupervised Hybrid-similarity Hadamard Hashing} (DU3H)~\cite{zhang2020deep}. DU3H first generates pseudo-labels through K-means clustering. Instead of adding a classification layer, DU3H utilizes Hadamard matrix to project pseudo-labels into Hamming space. This strategy is similar to CSQ~\cite{yuan2019central} but in unsupervised scenarios. Moreover, it generates a similarity structure for preserving pairwise similarity, which considers the confidence of different signals. This consideration of different confidence can also be seen in UDMSH~\cite{qin2020unsupervised} and DSAH~\cite{lin2021deep}. 
Lastly, a two-layer GCN is introduced to amplify the discrepancy of similarity signals to further guide the hash code learning. DU3H combines pointwise methods and pairwise methods in recent deep supervised learning domains, which helps to achieve significant improvement.} 

{\emph{Unsupervised Deep K-means Hashing} (UDKH)~\cite{dong2020unsupervised}. UDKH combines deep clustering with K-means. It first uses K-means clustering results to initialize the cluster labels. UDKH learns both hash codes and cluster labels in an alternative manner. Specifically, it first fixes clustering results and optimizes the hash codes as well as the hashing network under supervision. Then, it fixes the hash codes and leverages Discrete Proximal Linearized Minimization~\cite{shen2016fast} to derive the updated pseudo-labels. UDKH repeats the above steps until convergence. UDKH improves the quality of hash codes along with the pseudo-labels with progressive learning, achieving better performance compared with unsupervised methods using fixed pseudo-labels.}

\subsection{Prediction-free self-supervised learning-based Methods}\label{sec:ssm}

{The last class of deep unsupervised methods is prediction-free self-supervised learning-based methods. The early methods often impose several constraints on hash codes by minimizing regularization terms (i.e., the bit balance loss, the bit independence loss the quantization loss and transformation-invariant loss) ~\cite{erin2015deep,lin2016learning}. 
To extract more information through deep neural networks, several researchers introduce popular self-supervised techniques into deep unsupervised hashing, such as auto-encoder~\cite{do2016learning,dai2017stochastic,shen2020auto} and generative adversarial network~\cite{ghasedi2018unsupervised,song2018binary,zieba2018bingan}, etc. Recently, contrastive learning has shown promising performance in producing discriminative representations in various domains.  
Inspired by the fact that hash code is a specific form of representation, several methods involve contrastive learning into recent unsupervised hashing, which helps to get high-quality hash codes~\cite{luo2020cimon,qiu2021unsupervised,jang2021self}. These methods usually first transform each input $\bm{x}_i$ into two views ${\bm{x}}_i^{(1)}$ and ${\bm{x}}_i^{(2)}$. Then the hashing network projects them into two hash codes ${\bm{b}}_i^{(1)}$ and ${\bm{b}}_i^{(2)}$. Given the $\bm{\alpha} \star \bm{\beta}$ denotes the cosine similarity of two vectors, the network is trained by minimizing the loss for each batch as follows:
\begin{equation}\label{eq:cl}
\mathcal{L}_{CL}=-\frac{1}{2 N_B} \sum_{i=1}^{N_B}\left(\log \frac{e^{\bm{b}_{i}^{(1)} \star \bm{b}_{i}^{(2)} / \tau}}{Z_{i}^{(1)}}+\log \frac{e^{\bm{b}_{i}^{(1)} \star \bm{b}_{i}^{(2)} / \tau}}{Z_{i}^{(2)}}\right)
\end{equation}
where $\tau$ is a temperature parameter, $N_B$ is the batch size and $
Z_{i}^{(r)}=\sum_{j \neq i}\left(e^{\bm{b}^{(r)} \star \bm{b}_{j}^{(1)} / \tau}+e^{\bm{b}_i^{(r)} \star \bm{b}_j^{(2)} / \tau}\right)
$, $r=1 $ or $2$. Minimizing Eq. \ref{eq:cl} has the following strengths. First, since the numerator penalizes the difference in binary codes of samples under different views, it assists in the production of transformation-invariant binary codes. Second, since the denominator promotes to amply the distances between binary codes of different examples which facilities the binary codes to approach a uniform distribution in the Hamming space~\cite{pmlr-v119-wang20k}, it assists in optimizing the capacity of hash bits \cite{shen2018unsupervised}, preserving the most semantic information. Third, because contrastive learning demonstrates promising performance in various tasks including linear classification as well as clustering  \cite{he2020momentum,li2020contrastive}, it aids in developing high-quality binary codes for effective retrieval.
}

{\emph{Deep Hashing} (DH)~\cite{erin2015deep}. DH utilizes a deep hashing network and optimizes the parameters of the network with three criteria for the hash codes. Firstly, it minimizes a quantization loss by minimizing the gap between the network output and the learnt hash codes. Secondly, it minimizes the bit balance loss in Eq. \ref{eq:8} so that generated binary codes distribute evenly on each bit. Thirdly, it regularizes the weights of hashing network for independent hash codes. The parameters of the hashing network are updated by back-propagation based on the composite objective function. DH only imposes several constraints on hash codes without inferring similarity information from the training data, which results in limited performance.}

{\emph{DeepBit}~\cite{lin2016learning}. DeepBit utilizes a deep convolutional neural network as the backbone. It also minimizes the quantization loss as well as the bit balance loss. Differently, DeepBit enforce the hash codes invariant to image rotation. The rotation invariant loss is formulated as:
\begin{equation}
 \mathcal{L}(RI)= \sum_{i=1}^N\sum_{\theta=-R}^{R} exp(-\frac{\theta^2}{2})\left\|\bm{h}_i- \bm{h}_{i,\theta}\right\|^{2}
\end{equation}
where $\theta$ is the rotation angle and $\bm{h}_{i,\theta}$ denotes the network output from $\bm{x}_i$ with rotation $\theta$. This loss acts as a regularization term to enforce the hash codes invariant to certain transformations, which improves the performance compared with DH.}

{\emph{Unsupervised Triplet Hashing} (UTH)~\cite{huang2017unsupervised} UTH builds the triplets from the dataset, each of which contains an anchor example, a rotated example along with a random example. Afterward, the hashing network is optimized using the triplet inputs. The quantization loss and bit balance loss are also adopted for high-quality hash codes. The triplet loss compares the hash codes from different hash codes, which helps generate discriminative hash codes compared with the regularization loss in DeepBit. Hence, UTH performs better than DeepBit in various experiments.}

{\emph{HashGAN}~\cite{ghasedi2018unsupervised}. HashGAN contains three networks, i.e., a generator, a discriminator and a hashing network. The hashing network utilizes $L$ sigmoid function for final activation.
Its objective for real data contains four losses. It first minimizes the entropy of each bit, which is equivalent to a quantization loss. The other three terms enforce the bit balance, invariance to different transformations and bit independence. Similar to DSH-GAN, The discriminator is trained in an adversarial form. It also leverages the synthesized images by minimizing the distances between outputs of the hashing network and the inputs of the generator, which acts like an auto-encoder. Moreover, it encourages the generator to produce synthetic samples with similar statistics to real samples with $L_2$-norm loss. With GAN, HashGAN achieves better performance on both information retrieval and clustering tasks.}

{\emph{Stochastic Generative Hashing} (SGH) \cite{dai2017stochastic}. SGH proposes to utilize a generative manner to train the hashing network through the Minimum Description Length principle. In this manner, the obtained binary codes compress the whole dataset as much as possible. Specifically, it contains a generative network and an encoding network to build the mapping between inputs and binary codes from adverse directions. During optimization, it trains a variational auto-encoder to reconstruct the input using the least information in binary codes. SGH is a general framework, which can be degraded into ITQ~\cite{gong2012iterative} as well as Binary Autoencoder~\cite{carreira2015hashing}.}

{\emph{Unsupervised Hashing with Contrastive Information Bottleneck} (CIBHash)~\cite{qiu2021unsupervised}. CIBHash adapts contrastive learning in deep unsupervised hashing. It considers the outputs of the hashing network as a form of representation and minimizes the contrastive loss on the outputs. Specifically, CIBHash generates two views for each input, and minimizes the contrastive learning objective, i.e., Eq. \ref{eq:cl}. To estimate the gradient of hashing network with discrete stochastic variables, CIBHash leverages the straight-through gradient estimator~\cite{bengio2013estimating} and the gradients are transmitted entirely to the front layer similar to \cite{su2018greedy}. CIBHash also illustrates the objective with Information bottleneck theory with an improved model variant. From that moment on, contrastive learning has been shown an effective tool for deep unsupervised learning since then.} 

{\emph{Self-supervised Product Quantization} (SPQ)~\cite{jang2021self}. SPQ combines contrastive learning with deep quantization. The codewords and deep continuous representations are simultaneously optimized by contrasting individually augmented views in a cross manner. Specifically, for two views of each sample, i.e., ${\bm{x}}_i^{(1)}$ and ${\bm{x}}_i^{(2)}$, SPQ generates deep features ${\bm{z}}_i^{(1)}$ and ${\bm{z}}_i^{(2)}$ and employs codebooks in the quantization head to generate quantized features ${\hat{\bm{z}}}_i^{(1)}$ and ${\hat{\bm{z}}}_i^{(2)}$. Instead of comparing similarity between two visual descriptors or two quantized features, SPQ attempt to maximizes cross-similarity between the continuous representation from one perspective and the feature after product quantization from the other perspective. In formulation, 
\begin{equation}
\mathcal{L}_{SPQ}=-\frac{1}{2 N_B} \sum_{i=1}^{N_B}\left(\log \frac{e^{{\bm{z}}_i^{(1)} \star {\hat{\bm{z}}}_i^{(1)} / \tau}}{Z_{i}^{(2)}}+\log \frac{e^{\hat{\bm{z}}_i^{(1)} \star {{\bm{z}}}_i^{(2)} / \tau}}{Z_{i}^{(2)}}\right)
\end{equation}
where $Z_i^{(1)}=\sum_{j\neq i} e^{{\bm{z}}_i^{(1)} \star {\hat{\bm{z}}}_j^{(2)} / \tau}$ and $Z_i^{(2)}=\sum_{j\neq i} e^{\hat{\bm{z}}_i^{(2)} \star {{\bm{z}}}_j^{(1)} / \tau}$. With the cross contrastive learning strategy, both codewords and continuous representations are concurrently optimized to produce high-quality outputs for effective image retrieval.}

{\emph{Hashing via Structural and Intrinsic Similarity Learning} (HashSIM)~\cite{luo2022improve}. HashSIM utilizes contrastive learning for deep unsupervised hashing from a different view. For each batch, it stacks two views of binary codes into two distinct matrices $\bm{B}^{(1)}$ and $\bm{B}^{(2)} \in \mathbb{R}^{N_B\times L}$, and takes their column vectors as bit vectors $\{\bm{c}_l^{(r)}\}_{l=1}^L$, $r=1$ or $2$. Then, HashSIM develops a intrinsic similarity learning objective as follows:
\begin{equation}
\mathcal{L}_{HashSIM}=-\frac{1}{2L} \sum_{l=1}^{L}\left(\log \frac{e^{\bm{c}_{l}^{(1)} \star \bm{c}_{l}^{(2)} / \tau}}{Z_{i}^{(1)}}+\log \frac{e^{\bm{c}_{l}^{(1)} \star \bm{c}_{l}^{(2)} / \tau}}{Z_{l}^{(2)}}\right)
\end{equation}
where $
Z_{i}^{(r)}=\sum_{l' \neq l}\left(e^{\bm{c}_l^{(r)} \star \bm{c}_{l'}^{(1)} / \tau}+e^{\bm{c}_{l}^{(r)} \star \bm{c}_{l'}^{(2)} / \tau}\right)$. Due to the fact the numerator attempts to reduce the gap between each hash bit under distinct augmentations and the denominator attempts to enlarge the distance between distinct bits, minimizing this self-supervised objective helps produce robust and independent hash codes for effective image retrieval. }

\section{Related Important Topics}

\subsection{Semi-supervised Deep Hashing}

{Semi-supervised deep hashing simultaneously leverages the semantic information from both labeled samples and unlabeled samples, and a range of semi-supervised deep hashing models have been developed recently. Compared with supervised methods and unsupervised methods, these methods can typically overcome label scarcity in practical with limited performance degradation.
These methods usually incorporate semi-supervised techniques (e.g., pairwise pseudo-labeling~\cite{zhang2017ssdh,shi2020anchor,yan2017semi}, GAN~\cite{wang2018semi,hu2021adversarial} and transductive learning~\cite{shi2021transductive}) into deep semi-supervised hashing. Then the retrieval performance can benefit from abundant unlabeled images in the real world. Generally, semi-supervised deep hashing provides a cost-effective solution to practical applications with promising performance, which desires further study in large-scale scenarios. We then review these methods in detail. }

{\emph{Semi-Supervised Deep Hashing} (SSDH)~\cite{zhang2017ssdh}. SSDH minimizes the semi-supervised loss function containing three terms, i.e., a ranking term, a embedding term, as well as a pseudo-label term. Supervised ranking term leverages a triplet loss for labeled data. Then SSDH generates an online k-NN graph for all data, which guides pairwise similarity preserving of the hashing network. Moreover, in semi-supervised settings, it generates pseudo-labels which further guide the similarity preserving process. SSDH is the first to perform deep hashing in a semi-supervised fashion. }

{\emph{Deep Hashing with a Bipartite Graph} (BGDH)~\cite{yan2017semi}. BGDH builds a bipartite graph to uncover the latent semantic structure for unlabeled data. Different from the similarity graph in unsupervised hashing, its similarity structure is based on the relationships between labeled examples and unlabeled examples, resulting in a bipartite graph. Then BGDH utilizes the bipartite graph to guide hash code learning by pairwise similarity preserving. It also adopts the loss term in DPSH~\cite{li2016feature} for supervised learning. Through mining the relationship in deep feature space, BGDH utilizes unlabeled data in an appropriate manner and improves the performance.} 

{\emph{Semi-Supervised Generative Adversarial Hashing} (SSGAH)~\cite{wang2018semi}. SSGAH combines a Generative Adversarial Network with deep semi-supervised hashing. It contains a generative network, a discriminator and a deep hashing network. The generative network produces two synthetic images $\bm{x}_{syn}^p$ and $\bm{x}_{syn}^n$ for each real image $\bm{x}$ and the similarity between $\bm{x}$ and $\bm{x}_{syn}^p$ is larger than the similarity between $\bm{x}$ and $\bm{x}_{syn}^n$. In this way, SSGAH learns the distribution of triplet-wise semantic message from both labeled samples as well as unlabeled samples. The discriminator estimates the likelihood that each input is synthetic. The hashing network is optimized using a triplet loss with the incorporation of synthetic positive and negative images. SSGAH can produce hash codes which could sufficiently explore semantics in the datasets by training the framework using an adversarial manner.}

{\emph{Semi-supervised Deep Pairwise Hashing} (SSDPH)~\cite{shi2020anchor}. SSDPH chooses a variety of labeled anchors in the training set, and then uses the pairwise objective for preserving similarities between labeled samples. More importantly, it leverages the technique of temporal ensembling from semi-supervised learning for learning similarity information from unlabeled data. Specifically, it contains a teacher model and a student model. The teacher model provides supervised information to guide the similarity learning, which is then updated in an ensemble manner. SSDPH first combines deep hashing with semi-supervised techniques, which improves the retrieval performance in real-world applications. }

{\emph{Transductive Semi-supervised Deep Hashing} (TSDH)~\cite{shi2021transductive}. TSDH extends the traditional transductive learning principle into deep semi-supervised hashing, which treats pseudo-labels of unlabeled data as variables and optimizes them alternatively with the hashing network. To accomplish this, it adds a classification layer after producing hash codes. Moreover, it involves a pairwise loss for similarity preservation. Lastly, TSDH estimates the confidence of pseudo-labels by the proximity distance:
\begin{equation}
v_{i}=\sum_{\bm{z}_{j} \in \mathcal{N}\left(\bm{z}_{i}\right)}\left\|\bm{z}_{i}-\bm{z}_{j}\right\|_{2}
\end{equation}
\begin{equation}
r_{i}=1-\frac{v_{i}}{v_{\max }}, \quad v_{\max }=\max \left\{v_{1}, \ldots, v_{N}\right\}
\end{equation}
where $\bm{z}_i$ is the extracted features of $\bm{x}_i$ and $N(\bm{z}_i)$ denotes the k-nearest neighbor set of $\bm{z}_i$. In this manner, samples that reside in densely populated regions are assigned a high confidence level. In summary, TSDH utilizes the popular transductive learning technique to improve the retrieval performance of semi-supervised hashing.}

{\emph{Adversarial Binary Mutual Learning} (ACML)~\cite{hu2021adversarial}. ACML also integrates a Generative Adversarial Network into semi-supervised deep hashing. Specifically, it 
contains a discriminative network and a generation network to mould the relationships between inputs and binary codes from opposite views. Then, an adversarial network is trained to differentiate between real and fake pairs of samples and their hash codes. In this way, it can leverage unlabeled data to make the discriminative network and generation network mutually learn from each other. Moreover, it introduces a Weibull distribution for better similarity preserving. ACML combines a Generative Adversarial Network with deep semi-supervised hashing and shows promising retrieval performance.}

\subsection{Domain Adaptation Deep Hashing}
{The data in the domain of interest is likely to be insufficient in practice while the labeled samples from a separate but correlated domain is usually accessible. To sufficiently utilize the labeled samples from source domains, several domain adaptive hashing methods have been developed in recent years. These hashing methods usually combine similarity preserving techniques (e.g., pairwise~\cite{zhou2018transfer,venkateswara2017deep} and ranking-based similarity preserving~\cite{long2018deep}) in deep supervised hashing with domain adaptation techniques (e.g., discrepancy minimization~\cite{zhang2019optimal,huang2021domain}, adversarial learning~\cite{he2019one,long2018deep} and centroid alignment~\cite{venkateswara2017deep,he2019one}). Hence, their methods are quite flexible. However, the cross-domain retrieval performance of current hashing methods is still not satisfactory, which desires further exploration in the future. 
We then review these methods as follows. }

{\emph{Domain Adaptive Hashing} (DAH) ~\cite{venkateswara2017deep}. DAH contains three parts, i.e., a supervised hashing module for source data, an unsupervised hashing module for target data and a domain disparity reduction module. For source data, it minimizes the likelihood loss along with the quantization loss. For source data, it leverages the source output to generate the label distributions and then minimizes the entropy to ensure that the target outputs approximate source outputs from each category. Further, DAH reduces the domain difference between the source and target representations through the minimization of multi-kernel Maximum Mean Discrepancy. This work is the first to combine unsupervised domain adaptation with deep hashing and improves the efficiency for cross-domain image retrieval.}

{\emph{Domain Adaptive Hashing with Intersectant
Generative Adversarial Networks} (IGAN)~\cite{he2019one} Different from DAH, IGAN generates the pseudo-labels for target domains and then aligns the semantic centroid for all categories. Moreover, it leverages two generators to reconstruct images in two domains and the generators and the discriminators are updated using a GAN objective. IGAN improves the retrieval performance using GAN as well as centroid alignment, which are two common techniques in domain adaption.}

{\emph{Deep Domain Adaptation Hashing with Adversarial Learning} (DeDAHA)~\cite{long2018deep}. DeDAHA contains two different CNNs for learning image representations. An adversarial loss is enforced to explore the knowledge robust to different domains. Then DeDAHA utilizes a standard triplet loss to learn the hashing encoder. When the label annotations in target data are unavailable, DeDAHA leverages a multi-stage framework for unsupervised domain adaptation hashing. }

{\emph{Deep Transfer Hashing} (DPH)~\cite{zhou2018transfer} DPH first uses a neural network to extract deep features and then incorporates a deep transformation mapping network for domain adaptation. Then for effective transfer learning, DPH generates the similarity information based on the cosine similarity of deep features as well as the hash codes in source domains to guide hash code learning. DPH shows great generality utilizing the powerful representative capacity
of deep learning.}

{\emph{Optimal Projection Guided Transfer Hashing} (GTH)~\cite{zhang2019optimal}. GTH seeks for the maximum likelihood estimation solution to minimize the error matrix between two hash projections of target and source domains. In this way, GTH can produce domain-invariant hash projections for effective cross-domain image retrieval. However, GTH assumes that similar domains should have small discrepancies between hash projections, which may be not promised in most scenarios. }

{\emph{Domain Adaptation Preconceived Hashing} (DAPH)~\cite{huang2021domain}. DAPH first reduces the
distribution discrepancy across two domains through learning a transformation matrix to project the samples from different domains into a common space. Moreover, it involves a reconstruction constraint to release the information loss from the transformation. For effective hash code learning, it adds a quantization loss to project features into hash codes. The whole learning process is in an alternative manner for updating the transformation matrix, projection and binary codes. DAPH improves the performance for challenging cross-domain retrieval.}

\subsection{Multi-modal Deep Hashing}

Multimedia data has exploded in multiple modalities including text, audio, image, and video since the dawn of the information era and the fast expansion of the Internet. Multi-model deep hashing has arisen much interest in the field of deep hashing recently. These methods typically project multiple modalities of data into a shared Hamming space using deep neural networks for effective cross-modal retrieval.
The framework of multi-modal deep hashing methods is similar to general deep hashing methods except that the similarity information includes the intra-modal and inter-modal forms. However, each loss term characterizing the similarity information is similar to that in deep supervised hashing discussed above. Existing methods~\cite{wang2021survey} can also be categorized into supervised methods~\cite{jiang2017deep,cao2018cross,yang2017pairwise} and unsupervised methods~\cite{yu2021deep,hu2020creating,wang2021set}.
Cao et al.~\cite{cao2020review} give a detailed review for the multi-modal hashing methods that includes \cite{jiang2017deep,yang2017pairwise,li2018self,zhang2018attention,chen2018supervised1,ji2019deep,yao2019discrete,gu2019adversary,cao2019hybrid,ding2016cross,hu2018deep,cao2018cross,hu2021video,li2021task}.

\section{Evaluation protocols}
\subsection{Evaluation Metrics}
For deep hashing algorithms, the space cost only depends on the length of the hash codes, so the length is usually kept the same when comparing the performance of different algorithms. The search efficiency is measured by the average search time for a query, which mainly depends on the architecture of the neural networks. Besides, if the weighted Hamming distance is used, we cannot take advantage of bit operation for efficiency.

As discussed above, we usually use search accuracy to measure performance. The most popular matrices include Mean Average Precision, Recall, Precision as well as the precision-recall curve. 
\emph{Precision}: Precision is defined by the proportion of returned samples that share the common label with the query. The formula can be formulated as:
\begin{equation}
precision=\frac{TP}{TP+FP},
\end{equation}
where $TP$ denotes the number of returned samples that have a common label with the query and $FP$ denotes the number of returned samples that do not have a common label with the query. $precision$@$k$ means the total number of returned sample is $k$, i.e. $TP+FP=k$. 

\emph{Recall}: Recall is defined by the proportion of samples in the database that have a common label with the query that is retrieved. The formula can be formulated as:
\begin{equation}
recall=\frac{TP}{TP+FN},
\end{equation}
where $FN$ is the total number of samples in the database that have a common label with the query, including samples not retrieved. $recall$@$k$ means the total number of returned samples is $k$. 

\emph{Precision-recall curve}: The precision rate and recall rate in image retrieval are both influenced by $k$. The precision and recall rates of an approach are inversely proportional. As a result, we could create the precision-recall curve by altering $k$ and using the precision rate and recall rate, respectively.

\emph{Mean average precision} (MAP): When the recall rate varies between 0 and 1, the average accuracy can be computed by varying the precision rate. The sequence summation approach is used to compute the average accuracy in practical applications with discretion:
\begin{equation}
AP=\frac{1}{F}\sum_{k=1}^N precision\text{@}k\Delta \{T\text{@}k\},
\end{equation}
in which $\Delta \{T\text{@}k\}$ denotes the change in recall from item $k-1$ to $k$. {The} sum of $\Delta \{T\text{@}k\}$ is $F$ and the core idea of AP is to evaluate a ranked list by averaging the precision at each position. Afterward, MAP can be derived by taking the mean of the average precision of every query. In several works, MAP is calculated in terms of top K ranked retrieval results. Some researchers also calculate the MAP with Hamming Radius r, when only samples with distances not bigger than $r$ are considered. 

Alexandre Sablayrolles et al.~\cite{sablayrolles2017should} show that the above popular evaluation protocols for supervised hashing are not satisfactory because a trivial solution that encodes the output of a classifier significantly outperforms existing methods. Furthermore, they provide a novel evaluation protocol based on retrieval of unseen classes and transfer learning. However, if the design of hashing methods avoids using the encoding of the classifier, the above popular evaluation protocols are still effective generally. 

\subsection{Datasets}
The scales of regularly used assessment datasets range from small to large to extremely large. Single-label datasets and multi-label datasets are two types of datasets.

\textbf{MNIST} \cite{lecun1998gradient} is comprised of 60,000 training samples and 10,000 testing samples. {It is a single-labeled dataset, where the ten different classes represent different digits.} Each image is represented by 784-dimensional raw features and 10,000 features as the queries.

\textbf{CIFAR-10} \cite{krizhevsky2009learning} is comprised of 60,000 real-world images in ten distinct categorizations. It is a single-labeled dataset, where ten different categorizations imply airplanes, cars, birds, cats, deer, dogs, frogs, horses, ships, as well as trucks. These examples are identified with semantic labels utilized to assess the performance of various hashing approaches. 

\textbf{ImageNet} \cite{deng2009imagenet} is a large-scale dataset that consists of over 1.2 million images hand-annotated by the huge project to find out what objects are included. It is a single-labeled dataset, consisting of 1,000 categories such as "balloon" or "strawberry".

\textbf{NUS-WIDE} \cite{chua2009nus} is a well-known {multi-labeled} image dataset collected by a team from NUS. It consists of 269,648 examples with 5,018 unique tags. These samples are manually associated with some of the 81 concepts. Because images have typically over one label, two samples are treated as semantic similar if they share one common semantic label.

\textbf{MS COCO} \cite{lin2014microsoft} {is a popular multi-labeled datasets, consisting} of 82,783 training examples along with 40,504 validation examples, each of which is associated with part of the eighty categories. After removing examples without any class information, 122,218 samples can be obtained for evaluating the performance of hashing methods.

\subsection{Performance Analysis}
\begin{table}
\centering
\renewcommand{\arraystretch}{1.3}
\caption{MAP for different hashing methods on CIFAR-10 and NUS-WIDE.}\label{s:exp:1}
	\begin{tabular}{c|c|c|c|c|c|c|c|c}
		\hline
		& \multicolumn{4}{c|}{CIFAR-10} & \multicolumn{4}{c}{NUS-WIDE} \\
		\hline
		Method & 12bits & 24bits & 32bits &48bits & 12bits & 24bits & 32bits &48bits\\
		\hline
		CNNH \cite{xia2014supervised} & 0.439 &0.511& 0.509&0.522&0.611&0.618&0.625&0.608 \\
		DNNH \cite{lai2015simultaneous} & 0.552 &0.566& 0.558&0.581&0.674 &0.697 &0.713 &0.715\\
		 DHN \cite{zhu2016deep}&0.555 &0.594 &0.603 &0.621 &0.708 &0.735 &0.748 &0.758\\
		 DRSCH \cite{zhang2015bit} & 0.614&0.621&0.628&0.630&0.618&0.622&0.622&0.627\\
		DSCH \cite{zhang2015bit} & 0.608&0.613&0.617&0.619&0.591&0.597&0.610&0.608\\
		DSRH \cite{zhao2015deep} &0.608&0.610&0.617&0.617&0.609&0.617&0.621&0.630\\
		DSH-GAN~\cite{qiu2017deep} & 0.735 & 0.781 & 0.787 & 0.802 & 0.838 & 0.856 & 0.861 & 0.863 \\
		DTSH \cite{wang2016deep} &0.710 &0.750 &0.765 & 0.774&0.773 &0.808 &0.812 &0.824\\
		DPSH \cite{li2016feature} &0.713 &0.727 &0.744 & 0.757 &0.752 &0.790 &0.794 &0.812\\
		DSDH \cite{li2017deep} &0.740 &0.786 &0.801 &0.820 &0.776 &0.808 &0.820 &0.829\\
		DQN \cite{cao2016deep} & 0.554 &0.558 &0.564 &0.580 &0.768 &0.776 &0.783 &0.792\\
		DSH \cite{liu2016deep} & 0.644 &0.742 & 0.770 & 0.799& 0.712& 0.731&0.740&0.748\\
		DCEH \cite{wu2019deep} & 0.745 &0.788 & 0.802 & 0.806 & 0.781& 0.816&0.827&0.839\\
		DDSH \cite{jiang2018deep}& 0.753 &0.776&0.803&0.811&0.776&0.803&0.810&0.817\\
		DFH \cite{li2019push} &0.803&0.825&0.831&0.844&0.795&0.823&0.833&0.842\\
		Greedy Hash \cite{su2018greedy}&0.774&0.795&0.810&0.822& - & - &- & -\\
		MIHash \cite{cakir2019hashing} & 0.738&0.775&0.791&0.816&0.773&0.820&0.831&0.843\\
		HBMP \cite{cakir2018hashing} & 0.799&0.804&0.830&0.831&0.757&0.805&0.822&0.840\\
		VDSH \cite{zhang2016efficient} & 0.538&0.541&0.545&0.548&0.769&0.796&0.803&0.807\\
		NMLayer~\cite{fu2019neurons} & 0.786 & 0.813 & 0.821 & 0.828 & 0.801 & 0.824 & 0.832 & 0.840 \\
		HashNet \cite{cao2017hashnet} &0.685&0.707&0.705&0.705&0.770&0.802&0.806& 0.816\\
		AnDSH \cite{zhou2019angular} &0.754&0.780&0.786&0.795&0.780&0.808&0.815& 0.823\\
		DISH~\cite{zhang2018deep} & 0.738 &  0.792 & 0.822 & 0.841 & 0.781 & 0.823 & 0.837 &0.840\\
		SRE~\cite{zhang2019deep} & 0.771 & 0.817 & 0.839 & 0.858 & 0.801 & 0.833 & 0.849 & 0.861 \\
		MLDH~\cite{ma2018multi} & 0.805 & 0.825 & 0.829 & 0.832 & 0.800 & 0.828 & 0.832 & 0.835 \\
		\hline
		SDH \cite{shen2015supervised} & 0.285 & 0.329& 0.341 &0.356 &0.568&0.600& 0.608&0.638\\
		KSH \cite{liu2012supervised} & 0.303 &0.337 &0.346&0.356 & 0.556&0.572&0.581&0.588\\
		ITQ \cite{gong2012iterative} & 0.127 & 0.128 &0.126& 0.129 & 0.454 & 0.406 & 0.405 & 0.400\\
		\hline

	\end{tabular}
\end{table}
\begin{table}
\centering
\renewcommand{\arraystretch}{1.3}
\caption{MAP for different hashing methods on ImageNet and MS COCO.}\label{s:exp:2}
	\begin{tabular}{c|c|c|c|c|c|c}
		\hline
		& \multicolumn{3}{c|}{ImageNet} & \multicolumn{3}{c}{MS COCO} \\
		\hline
		Method & 16bits & 32bits & 64bits& 16bits & 32bits & 64bits \\
		\hline
		DBH~\cite{lin2015deep} & 0.350 & 0.379 & 0.406 & 0.602 & 0.639 & 0.658 \\
		DHN \cite{zhu2016deep}&0.311 & 0.472&0.573 & 0.677& 0.701 & 0.694 \\
		CNNH \cite{xia2014supervised} & 0.281 & 0.449 & 0.553 & 0.564 & 0.574 & 0.567 \\
		DNNH \cite{lai2015simultaneous} & 0.290  & 0.460 & 0.565 & 0.593 & 0.603 & 0.609\\
		DTSH \cite{wang2016deep} &0.442 & 0.528 & 0.581 & 0.699 & 0.732 & 0.753\\
	    HashNet \cite{cao2017hashnet} & 0.505 & 0.630 & 0.683 & 0.687 & 0.718 & 0.736 \\
	    SDH \cite{erin2015deep} & 0.584 & 0.649 & 0.664 & 0.671 & 0.710 & 0.733\\
	    DPSH~\cite{li2016feature} & 0.326 & 0.546 & 0.654 & 0.634 & 0.676 & 0.726\\
	    DSH~\cite{liu2016deep}  & 0.348 & 0.550 & 0.665 & - & - & - \\
	    HashGAN~\cite{cao2018hashgan}  & - & - & - & 0.687 & 0.718 & 0.736 \\
	    DCH~\cite{cao2018deep}  & 0.717 & 0.763 & 0.787 & 0.700 & 0.691 & 0.680\\
	    HashMI \cite{cakir2019hashing} & 0.569 & 0.661 & 0.694 & - & - & - \\
		Greedy Hash~\cite{su2018greedy} & 0.570 & 0.639 & 0.659 & 0.677 & 0.722 & 0.740 \\
		JMLH~\cite{shen2019embarrassingly} & 0.517 & 0.621 & 0.662 & 0.689 & 0.733 & 0.758\\
		DPN~\cite{fan} & 0.592 & 0.670 & 0.703 & 0.668 & 0.721 & 0.752 \\
		CSQ~\cite{yuan2019central} & 0.717 & 0.763 & 0.804 & 0.742 & 0.806 & 0.829 \\
		OrthHash~\cite{hoe2021one} & 0.614 & 0.681 & 0.709 & 0.708 & 0.762 & 0.785 \\
		DSEH~\cite{li2018deep} & 0.715 & 0.753 & 0.760 & 0.735 & 0.773 & 0.781 \\
		
		PSLDH~\cite{tu2021partial} & 0.734 & 0.792 & 0.817 & 0.782 & 0.835 & 0.853 \\

		\hline
		SDH \cite{shen2015supervised} & 0.298 & 0.455 & 0.585 & 0.554 & 0.564 & 0.579 \\
		KSH \cite{liu2012supervised} & 0.159 & 0.297 & 0.394 & 0.521 & 0.534 & 0.536 \\
		ITQ-CCA \cite{gong2012iterative} & 0.265 & 0.436 & 0.576 & 0.565 & 0.562 & 0.501 \\
		ITQ \cite{gong2012iterative} & 0.325 & 0.462 & 0.552 & 0.581 & 0.624 & 0.657 \\
		BRE~\cite{kulis2009learning} & 0.062 & 0.252 & 0.357 & 0.592 & 0.622 & 0.633 \\
		SH~\cite{weiss2009spectral} & 0.206 & 0.328 & 0.419 & 0.495 & 0.509 & 0.510 \\
		LSH~\cite{gionis1999similarity} & 0.100 & 0.235 & 0.359 & 0.459 & 0.485 & 0.584 \\
		\hline

	\end{tabular}
\end{table}

\subsubsection{Performance Comparison of Deep Supervised Methods}\label{exp}
We present the results of some representative deep supervised hashing and quantization algorithms over CIFAR-10, NUS-WIDE, ImageNet and MS COCO. For CIFAR-10, 100 images are selected randomly per class (resulting in 1,000 images totally) as queries and the rest of samples are adopted as the database. 500 samples per class (resulting in 5,000 samples totally) make up the training set. For NUS-WIDE, a subset of 195,834 samples that correspond to the 21 most frequent labels are picked. Afterward, 100 samples per class (resulting in 2,100 samples totally) are picked as queries and the remaining samples make up the retrieval set. 500 images per class (resulting in 10,500 images totally) are sampled as the training set. For ImageNet, 100 categories are randomly selected. The samples associated with these categories in the training set make up the database, and the samples in the validation set are utilized as queries. 100 images per category are selected from the database for training. For MS COCO, 5,000 samples are used as queries and the rest are used as the database. 10,000 samples from the database are selected for training.

Note that for the various experimental settings, most of the experimental results are not shown in this summary in detail. 
The representative compared results of hashing methods are shown in Table \ref{s:exp:1} and \ref{s:exp:2}.
From the results, there are several observations as follows: 
\begin{itemize}
    \item Deep supervised hashing greatly outperforms traditional hashing methods (SDH and KSH) overall, {validating the strong representation-learning capacity of deep learning.}
    \item Similarity information is necessary for deep hashing. For deep supervised hashing methods in the early period (i.e., before 2016), hash codes are mostly obtained by transferring classification models without supervised similarity information while the methods with pairwise and {ranking} information outperform them.
    \item Label information helps to increase the performance of deep hashing. This point can be shown from the fact that DSDH outperforms DPSH evidently and the superiority of LabNet. {Moreover, several pointwise methods (CSQ, OrthHash and PSLDH) show comparable performance recently by mapping the labels into Hamming space, achieving impressive performance on large-scale datasets.}
    \item Several skills including regularization term, bit balance, ensemble learning and bit independence help to obtain accurate and robust performance, which can be seen from ablation studies in some papers~\cite{do2016learning}.
    \item Although supervised hashing methods have achieved remarkable performance, they are difficult to be applied in practice since large-scale data annotations are unaffordable. To address this problem, deep learning-based unsupervised methods provide a cost-effective solution for more practical applications.
    
\end{itemize}

\begin{table}
\centering
\renewcommand{\arraystretch}{1.3}
\caption{MAP for different unsupervised methods on CIFAR-10, NUS-WIDE and MS COCO.}\label{un:exp}
	\begin{tabular}{c|c|c|c|c|c|c|c|c|c}
		\hline
		& \multicolumn{3}{c|}{CIFAR-10} & \multicolumn{3}{c|}{NUS-WIDE} &  \multicolumn{3}{c}{MS COCO} \\
		\hline
		Method & 16bits & 32bits & 64bits& 16bits & 32bits & 64bits& 16bits & 32bits & 64bits \\
		\hline
		DeepBit~\cite{lin2016learning} & 0.194 & 0.249 & 0.277 & 0.392 & 0.403 & 0.429 & 0.399 & 0.410 & 0.475 \\
		SGH~\cite{dai2017stochastic} & 0.435 & 0.437 & 0.433 & 0.593 & 0.590 & 0.607 & 0.594 & 0.610 & 0.618  \\
		BGAN~\cite{song2018binary} & 0.525 & 0.531 & 0.562 & 0.684 & 0.714 & 0.730 & 0.645 & 0.682 & 0.707\\
		BinGAN~\cite{zieba2018bingan} & 0.476 & 0.512 & 0.520 & 0.654& 0.709 & 0.713 & 0.651 & 0.673 & 0.696 \\
		Greedy Hash~\cite{su2018greedy} & 0.448 & 0.473 & 0.501 & 0.633 & 0.691 & 0.731& 0.582 & 0.668 & 0.710 \\
		HashGAN~\cite{ghasedi2018unsupervised}  & 0.447 & 0.463 & 0.481 & - & - & -  & - & - & - \\
		UH-BDNN~\cite{do2016learning}  & 0.301 & 0.309 & 0.312 & - & - & -  & - & - & - \\
		UTH~\cite{lin2016learning} & 0.287 & 0.307 & 0.324 & 0.450 & 0.495 & 0.549 & 0.438 & 0.465 & 0.508 \\
		SSDH~\cite{yang2018semantic} & - & - & - & 0.580 & 0.593 & 0.610 & 0.540 & 0.566 & 0.593\\
		DistillHash~\cite{yang2019distillhash} & 0.285 & 0.294 & 0.308 & 0.627 & 0.656 & 0.671 & 0.546 & 0.566 & 0.593\\
	    MLS$^3$RUDH~\cite{tu2020mls3rduh} & - & - & - & 0.713 & 0.727 & 0.750 & 0.607 & 0.622 & 0.641 \\
	    TBH~\cite{shen2020auto} & 0.532 & 0.573 & 0.578 & 0.717 & 0.725 & 0.735 & 0.706 & 0.735 & 0.722\\
	    
	    GLC~\cite{luo2021deep} & - & - & - & 0.759 & 0.772  & 0.783  & 0.715   & 0.723  & 0.731 \\ 
	    DVB~\cite{shen2019unsupervised} & 0.403 & 0.422 & 0.446 & 0.604 & 0.632 & 0.665& 0.570 & 0.629 & 0.623 \\
	    CIBHash~\cite{qiu2021unsupervised} & 0.590 & 0.622 & 0.641 & 0.790 & 0.807 & 0.815 & 0.737 & 0.760 & 0.775 \\
	    DATE~\cite{luo2021mm} & 0.577 & 0.629 & 0.647 & 0.793 & 0.809 & 0.815 & - & - & - \\

	    \hline
	    ITQ~\cite{gong2012iterative} & 0.305 & 0.325 & 0.349 & 0.627 & 0.645 & 0.664 & 0.598 & 0.624 &0.648 \\
	    AGH~\cite{liu2011hashing} & 0.333 & 0.357 & 0.358 & 0.592 & 0.615 & 0.616 & 0.596 & 0.625 & 0.631 \\
	    DGH~\cite{liu2014discrete} & 0.335 & 0.353 & 0.361 & 0.572 & 0.607 & 0.627 & 0.613 & 0.631 & 0.638 \\
	   \hline

	\end{tabular}
\end{table}

\subsubsection{Performance Comparison of Deep Unsupervised Methods}

{This part presents the results of representative deep unsupervised hashing approaches over CIFAR-10, NUS-WIDE and MS COCO. We follows the setting in prior works~\cite{qiu2021unsupervised,shen2020auto,luo2021mm}. The dataset splits for training, testing and database are the same as Sec. \ref{exp}. Part of records are quoted from \cite{qiu2021unsupervised,shen2022learning,shen2020auto}.}

{The compared results are shown in Table \ref{un:exp}. From the results, we have the following observations:
\begin{itemize}
    \item Deep unsupervised hashing methods generally perform better than the traditional approaches (ITQ, AGH and DGH), suggesting that the powerful representation learning capacity of deep learning is beneficial to the retrieval performance of generated binary codes.
    \item The methods that only adopt regularization terms (DeepBits and UTH) obtain poor results among compared methods, demonstrating that the exploration of semantic information is indispensable for discriminative hash codes. 
    \item The methods that explore more accurate similarity structures (DATE and TBH) outperform early approaches that obtain similarity structure in a coarse manner (SSDH and DistillHash). The potential reason is that false similarity signals will result in error propagation during subsequent hash code learning, implying suboptimal performance.
    \item The methods utilizing contrastive learning (CIBHash and DATE) achieve superb performance among compared methods, which implies that contrastive learning is an effective tool for discriminative hash code learning. As the research moves along, deep unsupervised learning methods can even outperform part of deep supervised methods, which is really inspiring. 
\end{itemize}}

\begin{figure*}[h]
    \centering
    \includegraphics[width=0.6\textwidth,keepaspectratio=true]{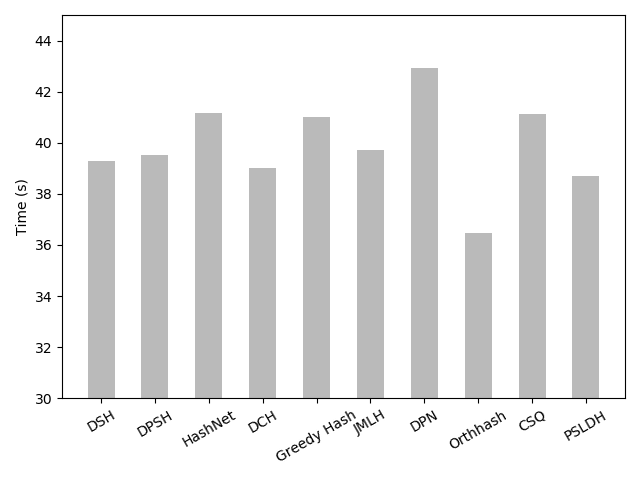}
    \caption{{Computational time cost of different hashing methods.}}\label{time}
\end{figure*}

\subsection{Training Time Cost}

In this part, we investigate the training efficiency of different deep hashing methods. Ten representative methods are selected. These methods are parameterized by different network backbones (e.g., AlexNet and VGG-F) and these backbones could bring in a larger difference of computational cost in training and inference compared with core hashing techniques. Hence,
all the hashing networks are parameterized by VGG-F and trained on a single NVIDIA GeForce GTX TITAN X GPU for fair comparison of the efficiency. 
In Fig. \ref{time}, we report the running time of each epoch during the training phase of different compared approaches. From the results, we have the following observations. First, the efficiency difference between these methods is limited. The potential reason is that the computational cost for hashing methods mainly depends on the forwarding and back propagation of the network backbone. The specific optimization manners have limited impacts on the computational cost. Second, OrthHash is the most effective among different methods, which is because that OrthHash only leverages one brief objective during optimization.

\section{Conclusion}
In this survey, we present a comprehensive review of the papers on deep hashing, including deep supervised hashing, deep unsupervised hashing and other related topics.  
Based on how measuring the similarities of hash codes, we divide deep supervised hashing methods into four categories: pairwise methods, ranking-based methods, pointwise methods and quantization. In addition, we categorize deep unsupervised hashing into three classes based on semantics learning manners, i.e., reconstruction-based methods, pseudo-label-based methods and prediction-free self-supervised learning-based methods. 
We also explore three important topics including semi-supervised deep hashing, domain adaption deep hashing and multi-modal deep hashing. We observe that the existing deep hashing methods mainly focus on the public datasets designed for classification and detection, which do not fully address the nearest neighbor search problem. Future works could attempt to combine downstream approximate nearest neighbor search algorithms to design specific deep hashing methods. In this way, researchers will propose more practical deep hashing methods for real-world applications. Further, cutting-edge deep neural network techniques and representation learning techniques will be integrated into deep hashing and promote the development of large-scale image retrieval.  
\begin{acks}
This work was supported by the National Key Research and Development Program of China (2021YFF1200902) and the National Natural Science
Foundation of China (31871342). We also thank Zeyu Ma, Huasong Zhong and Xiaokang Chen who discussed with us and provided instructive suggestions.
\end{acks}

\bibliographystyle{ACM-Reference-Format}
\bibliography{ms.bib}


\
\end{document}